\documentclass[sigconf]{acmart}

\usepackage{balance}

\acmSubmissionID{fp435}

\usepackage{subcaption}
\usepackage[linesnumbered,ruled]{algorithm2e}
\makeatletter
\DeclareRobustCommand\onedot{\futurelet\@let@token\@onedot}
\def\@onedot{\ifx\@let@token.\else.\null\fi\xspace}

\def\eg{\emph{e.g}\onedot}

\def\etal{\emph{et al}\onedot}
\makeatother

\setcopyright{acmlicensed}

\begin{document}

\fancyhead{}

\title{GP-GAN: Towards Realistic High-Resolution Image Blending}

%
\author{Huikai Wu}
\email{huikai.wu@nlpr.ia.ac.cn}
\orcid{}
\affiliation{
  \institution{CRISE, CASIA, and UCAS}
}
\author{Shuai Zheng}
\email{szheng@robots.ox.ac.uk}
\affiliation{
  \institution{University of Oxford}
}
\author{Junge Zhang}
\email{jgzhang@nlpr.ia.ac.cn}
\affiliation{
  \institution{CRISE, CASIA, and UCAS}
}
\author{Kaiqi Huang}
\email{kaiqi.huang@nlpr.ia.ac.cn}
\additionalaffiliation{CAS Center for Excellence in Brain Science and Intelligence Technology}
\affiliation{
  \institution{CRISE, CASIA, and UCAS}
}

%

%
\begin{abstract}
It is common but challenging to address high-resolution image blending in the automatic photo editing application. In this paper, we would like to focus on solving the problem of high-resolution image blending, where the composite images are provided. We propose a framework called Gaussian-Poisson Generative Adversarial Network (GP-GAN) to leverage the strengths of the classical gradient-based approach and Generative Adversarial Networks. To the best of our knowledge, it's the first work that explores the capability of GANs in high-resolution image blending task. Concretely, we propose Gaussian-Poisson Equation to formulate the high-resolution image blending problem, which is a joint optimization constrained by the gradient and color information. Inspired by the prior works, we obtain gradient information via applying gradient filters. To generate the color information, we propose a Blending GAN to learn the mapping between the composite images and the well-blended ones. Compared to the alternative methods, our approach can deliver high-resolution, realistic images with fewer bleedings and unpleasant artifacts. Experiments confirm that our approach achieves the state-of-the-art performance on Transient Attributes dataset. A user study on Amazon Mechanical Turk finds that the majority of workers are in favor of the proposed method. The source code is available in \url{https://github.com/wuhuikai/GP-GAN}, and there's also an online demo in \url{http://wuhuikai.me/DeepJS}.
\end{abstract}

\copyrightyear{2019} 
\acmYear{2019} 
\acmConference[MM '19]{Proceedings of the 27th ACM International Conference on Multimedia}{October 21--25, 2019}{Nice, France}
\acmBooktitle{Proceedings of the 27th ACM International Conference on Multimedia (MM '19), October 21--25, 2019, Nice, France}
\acmPrice{15.00}
\acmDOI{10.1145/3343031.3350944}
\acmISBN{978-1-4503-6889-6/19/10}

%
%


%

%
\begin{teaserfigure}
	\begin{subfigure}{0.24\linewidth}
		\includegraphics[width=\linewidth]{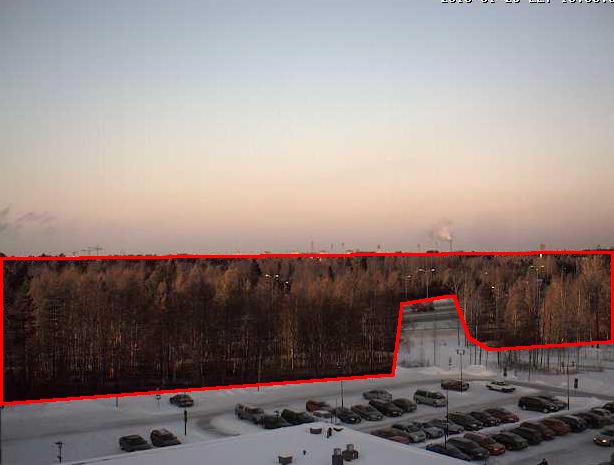}
		\caption{Copy-and-Paste Input}	\label{fig:intro_a}
	\end{subfigure}
	\hspace*{\fill}
	\begin{subfigure}{0.24\linewidth}
		\includegraphics[width=\linewidth]{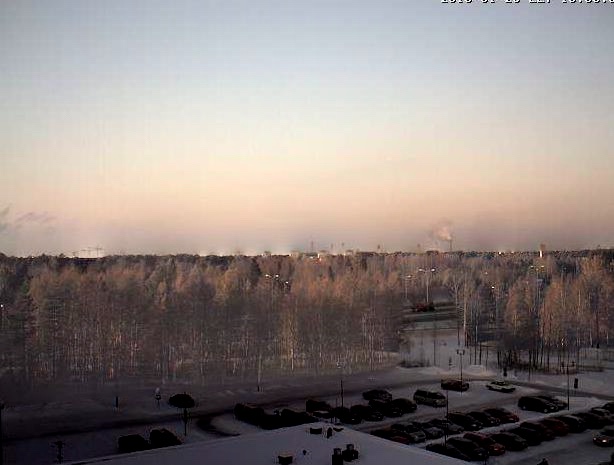}
		\caption{MPB~\cite{tanaka2012seamless}} \label{fig:intro_b}
	\end{subfigure}
	\hspace*{\fill}
	\begin{subfigure}{0.24\linewidth}
		\includegraphics[width=\linewidth]{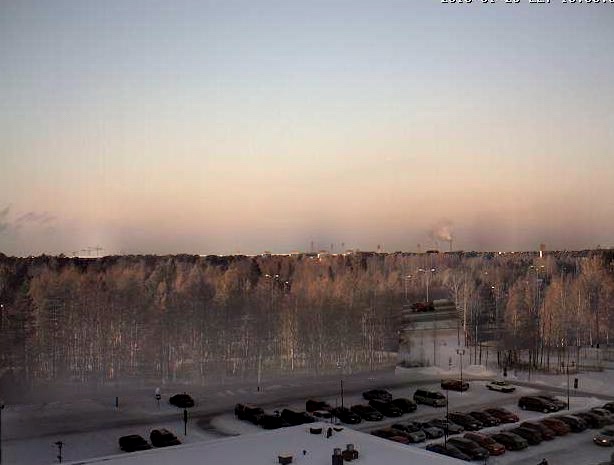}
		\caption{MSB~\cite{szeliski2011fast}} \label{fig:intro_c}
	\end{subfigure}
	\hspace*{\fill}
	\begin{subfigure}{0.24\linewidth}
		\includegraphics[width=\linewidth]{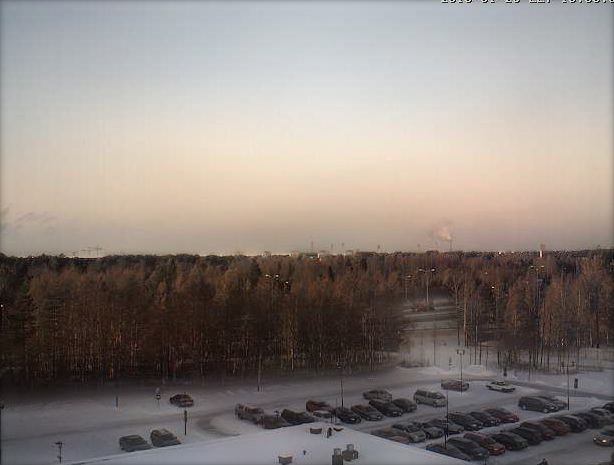}
		\caption{Ours} \label{fig:intro_d}
	\end{subfigure}
	\caption{Qualitative illustration of high-resolution image blending. (a) shows the composite copy-and-paste image, where the inserted object is circled out by the red polygon. Our approach (d) produces an image with better quality than those from the alternatives (b) and (c) in terms of illumination, spatial, and color consistencies. Best viewed in color.}
	\label{fig:intro}
	\Description{}
\end{teaserfigure}

%
\maketitle

\section{Introduction}
\begin{figure*}[h]
	\centering
	\includegraphics[width=0.9\linewidth]{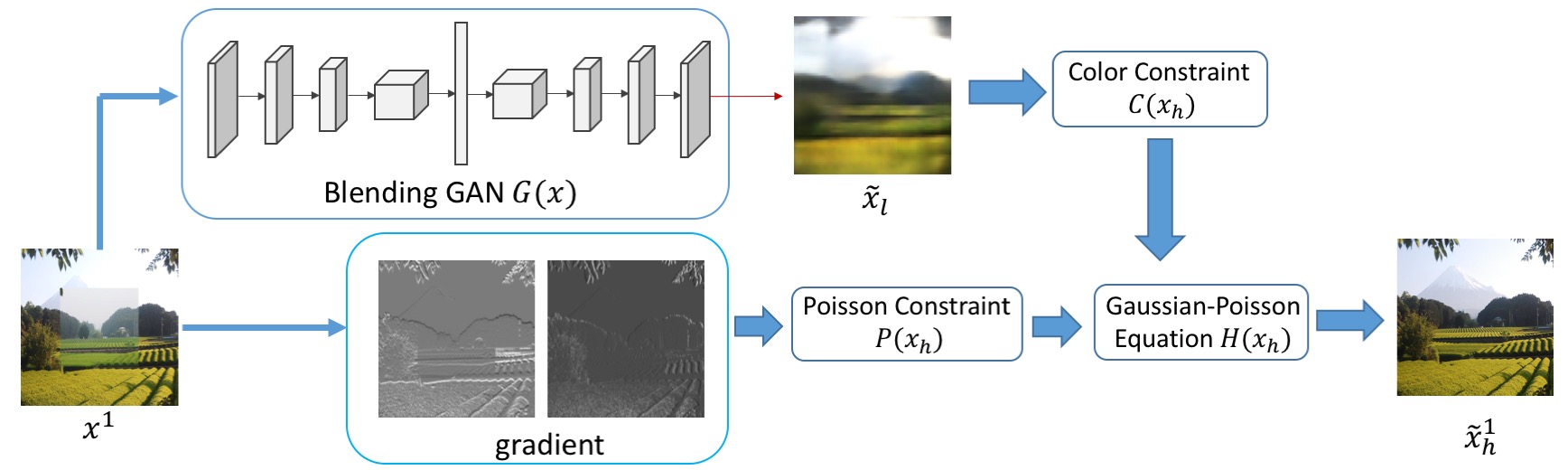}
	\caption{Framework Overview of GP-GAN. Given a composite image $x$, the low-resolution realistic image $\tilde{x}_l$ is first generated by Blending GAN $G(x)$ with $x^1$ as the input, where $x^1$ is the coarsest scale in the Laplacian pyramid of $x$. Then we optimize the Gaussian-Poison Equation $H(x_h)$ constrained by $C(x_h)$ and $P(x_h)$ with the closed-form solution, resulting in $\tilde{x}_h^1$ that contains rich details. We then upsample $\tilde{x}_h^1$ as the next $\tilde{x}_l$ and optimize the Gaussian-Poisson Equation at a finer scale in the pyramid of $x$. Best viewed in color.}
	\label{fig:framework}
\end{figure*}
Technologies such as PhotoShop make it much easier to edit an image than before. However, image editing still requires talents. For example, photos composited by expert users remain far better than the ones from newcomers. As the camera technologies improve, the high-resolution image makes photo editing becomes even more challenging. We want to bridge the talent gap between expert users and beginners on image editing. Mainly, we aim at addressing the problem of high-resolution image blending, which focuses on generating realistic high-resolution images given the composite ones.
As shown in Figure~\ref{fig:intro}, users insert an object in the background image (Figure~\ref{fig:intro_a}) and want to make it more realistic.
Most users would often have high expectation on the quality of the generated images.
If the algorithm produces images like Figure~\ref{fig:intro_b} or \ref{fig:intro_c}, users will give up the solution after their first few tries.

To generate well-blended images, Perez \etal, Tanaka \etal, and Szeliski \etal~\cite{perez2003poisson,tanaka2012seamless,szeliski2011fast} propose the classic gradient-based methods, which enable a smooth transition and reduce the color/illumination differences between foreground and background.
Among these solutions, Poisson image editing~\cite{perez2003poisson} is the most widely used method, which firstly produces a gradient vector field based on the gradients of the composite image and then recovers the blended image from this gradient vector field by addressing a Poisson equation.
Such methods are good at generating high-resolution results with rich details and textures.
However, the generated images tend to be unrealistic, which contain various kinds of artifacts.
Because the traditional gradient-based methods usually have strong assumptions about the distribution of realistic images based on human priors.

Recent researches have achieved significant progress in modeling the distribution of realistic images with the rise of Generative Adversarial Networks (GANs)~\cite{goodfellow2014generative, Denton/nips2015, Radford/iclr2016, arjovsky2017wasserstein}.
Concretely, GANs provide a framework for estimating the distribution of natural images via simultaneously training a generator and a discriminator in a zero-sum game.
The generator can produce natural images after training. 
Mirza~\etal~\cite{Mirza/arxiv/2014,Isola/cvpr2017} generalize the idea to a condition setting, which expands the usage of GANs into image-to-image applications like image inpainting~\cite{pathak2016context}.
Inspired by the success of GANs in generating realistic images, we propose to employ GANs for overcoming the disadvantages of gradient-based image blending algorithms.
Compared to these methods, GANs are much better at modeling the distribution of realistic images.
However, it usually takes lots of computation and memory resources to generate high-resolution images with rich details and textures.

We develop a novel framework named GP-GAN to combine the strength of GANs and gradient-based image blending methods, as shown in Figure~\ref{fig:framework}, which consists of two phases. In phase one, a low-resolution realistic image is generated based on the proposed Blending GAN. In phase two, we solve the proposed Gaussian-Poisson Equation based on the gradient vector field and the generated image in phase one fashioned by the Laplacian pyramid. This framework allows us to achieve high-resolution and realistic images, as shown in Figure~\ref{fig:intro_d}, which outperforms all the baseline methods. Our main contributions are four folds, which are summarized as follows:
\begin{itemize}
  \item We develop a framework GP-GAN for high-resolution image blending that takes advantages of both GANs and gradient-based image blending methods. To the best of our knowledge, it is the first work that explores the capability of GANs in high-resolution image blending task.
  \item We propose a network called Blending GAN for generating low-resolution realistic images.
  \item We propose the Gaussian-Poisson Equation for combining gradient information and color information.
  \item We also conduct a systematic evolution of the proposed approach based on both benchmark experiments and user studies on Amazon Mechanic Turk, which shows that our method outperforms all the baselines and achieves the state-of-the-art performance.
\end{itemize}

\section{Related Work}
We briefly review the relevant works from the classical image blending approaches to generative adversarial networks and conditional generative adversarial networks. We also discuss the difference between our work and the others.

\subsection{Image Blending}
The goal of classical image blending approaches is to improve the spatial and color consistencies between the source and target images. One way~\cite{Gracias/bmvc2006} is to apply the dense image matching approach to copy and paste the corresponding pixels. However, this method would not work when there are significant differences between the source and target images. The other way is to make the transition as smooth as possible for hiding artifacts in the composite images. Alpha blending~\cite{Uyttendaele/cvpr2001} is the simplest and fastest method, but it blurs the fine details when there are some registration errors between the source and target images. Alternatively, \cite{Levin/cvpr2001,Uyttendaele/cvpr2001,Fattal/2002,Agarwala/tog2004,Jia/tog/2006,Kazhdan/2008,Szeliski/2008} address this problem in the gradient domain. Our work is different from these gradient-based approaches in that we introduce GANs to generate a low-resolution realistic image as the color constraint, resulting in a more natural composite image.
\cite{xue2012understanding,zhu2015learning,tsai2017deep} also address a similar task to ours.
However, they focus on adjusting the color and illumination of the inserted object, requiring an accurate segmentation mask.
Differently, our method aims at making a smooth transition around the edges of the source and target images as well as reducing the color and illumination differences.
Thus, a well-blended image can be generated by our method, although the segmentation mask of the inserted object is coarse.

\subsection{Generative Adversarial Networks} Generative Adversarial Networks (GANs)~\cite{goodfellow2014generative} are first introduced to address the problem of generating realistic images. The main idea of GANs is a zero-sum game between learning a generator and a discriminator. The generator tries to produce more realistic images from random noises, while the discriminator aims to distinguish generated images from the real ones. Although the original method works for creating digital images from MNIST dataset, some generated images are noisy and incomprehensible. Denton~\etal~\cite{Denton/nips2015} improve the quality of the generated images by expanding GANs with a Laplacian pyramid implementation, but it does not work well for the images containing objects looking wobbly. Gregor~\etal~\cite{Gregor/nips2015} and Dosovitskiy~\etal~\cite{dosovitskiy2016generating} achieve successes in generating natural images; however, they do not leverage the generators for supervised learning. Radfor~\etal~\cite{Radford/iclr2016} achieve further improvement with deeper convolutional network architecture, while Zhang~\etal~\cite{Zhang/arxiv2016} stack two generators to progressively render more realistic images. InfoGAN~\cite{Chen/NIPS2016} learns a more interpretable latent representation. Salimans~\etal~\cite{Salimans/nips2016} reveal several tricks in training GANs. Arjovsky~\etal~\cite{arjovsky2017wasserstein} introduce an alternative training method Wasserstein GAN, which relaxes the GAN training requirement of balancing the discriminator and generator. However, existing GANs still do not work well for the image editing applications in that the generated results are not high-resolution and realistic yet.

\subsection{Conditional GANs} Our work is also related to conditional GANs~\cite{Mirza/arxiv/2014}, which aims to apply GANs in a conditional setting. There are several works along this research direction.
Previous works apply conditional GANs to discrete labels~\cite{Mirza/arxiv/2014}, text~\cite{Reed/icml2016}, image inpainting~\cite{pathak2016context}, image prediction from a normal map~\cite{Wang/eccv2016}, image manipulation guided by user constraints~\cite{zhu2016generative}, product photo generation~\cite{Yoo/eccv2016}, style transfer~\cite{Li/eccv2016}, and image-to-image translation~\cite{Isola/cvpr2017}. Different from previous works, we use an improved adversarial loss and discriminator for training the proposed Blending GAN. We also propose the Gaussian-Poisson Equation to produce high-resolution images.

\section{The Approach}
\label{section:approch}
In this section, we first introduce the task of image blending formally. We then present the framework of our Gaussian-Poisson Generative Adversarial Network (GP-GAN).
\begin{figure*}[h]
    \centering
    \includegraphics[width=0.95\linewidth]{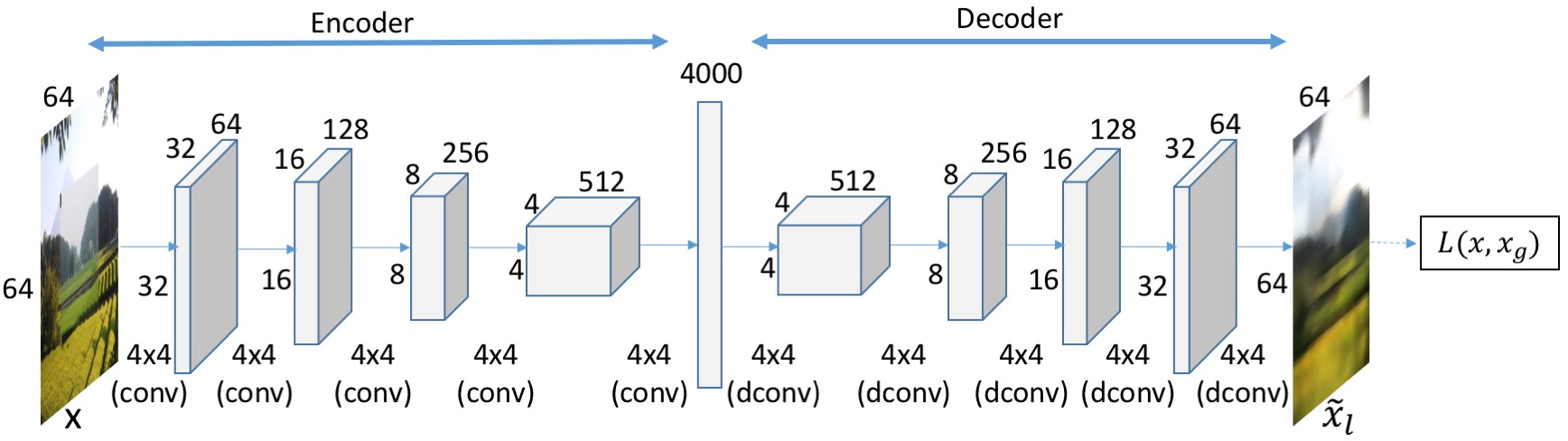}

    \caption{Network architecture of Blending GAN $G(x)$.}
    \label{fig:encoder-decoder}
\end{figure*}
\begin{figure}[h]
    \centering
    \includegraphics[width=\linewidth]{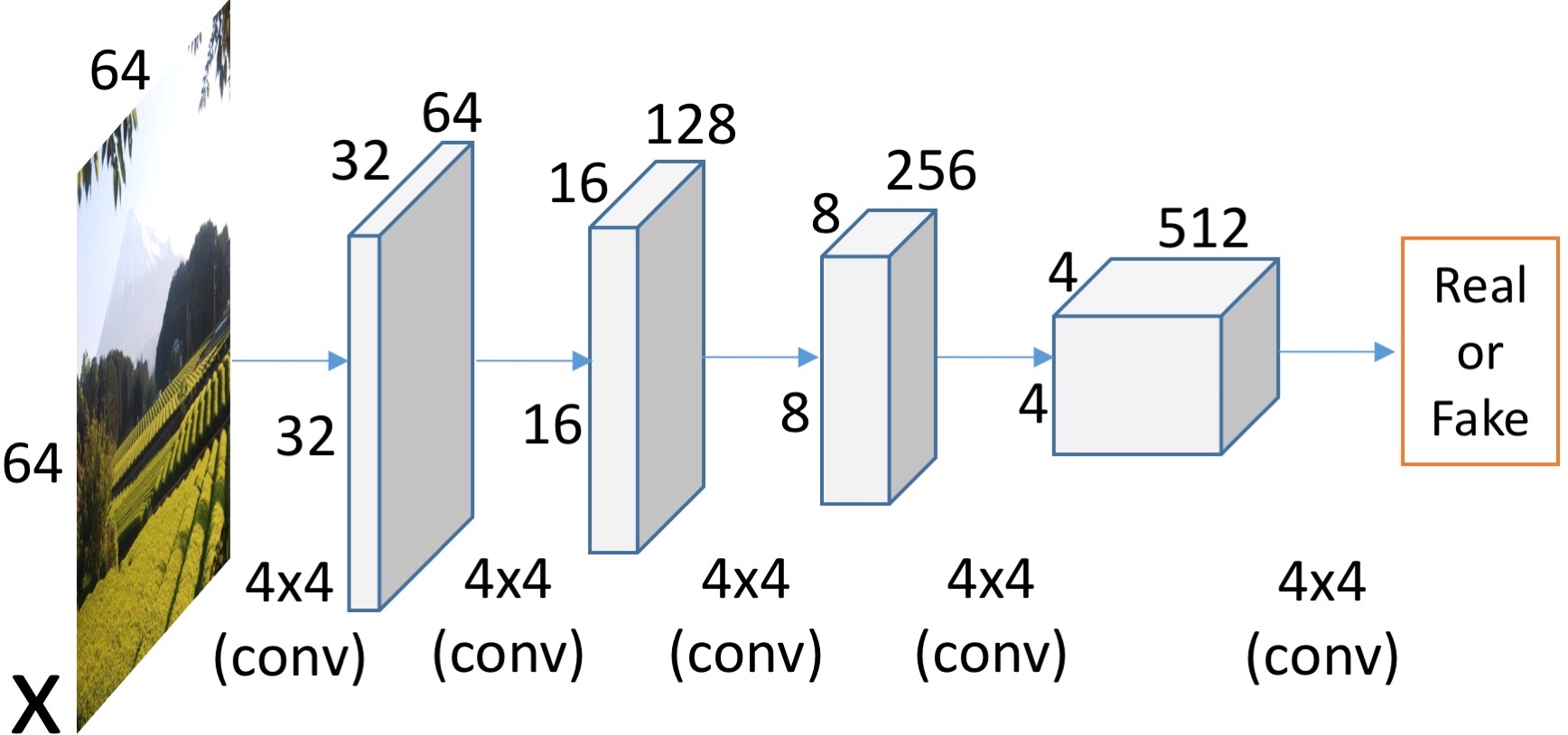}
    \caption{The architecture for the discriminator of Blending GAN.}
\label{fig:discriminator}
\end{figure}

\begin{figure}[h]
    \centering
    \includegraphics[width=\linewidth]{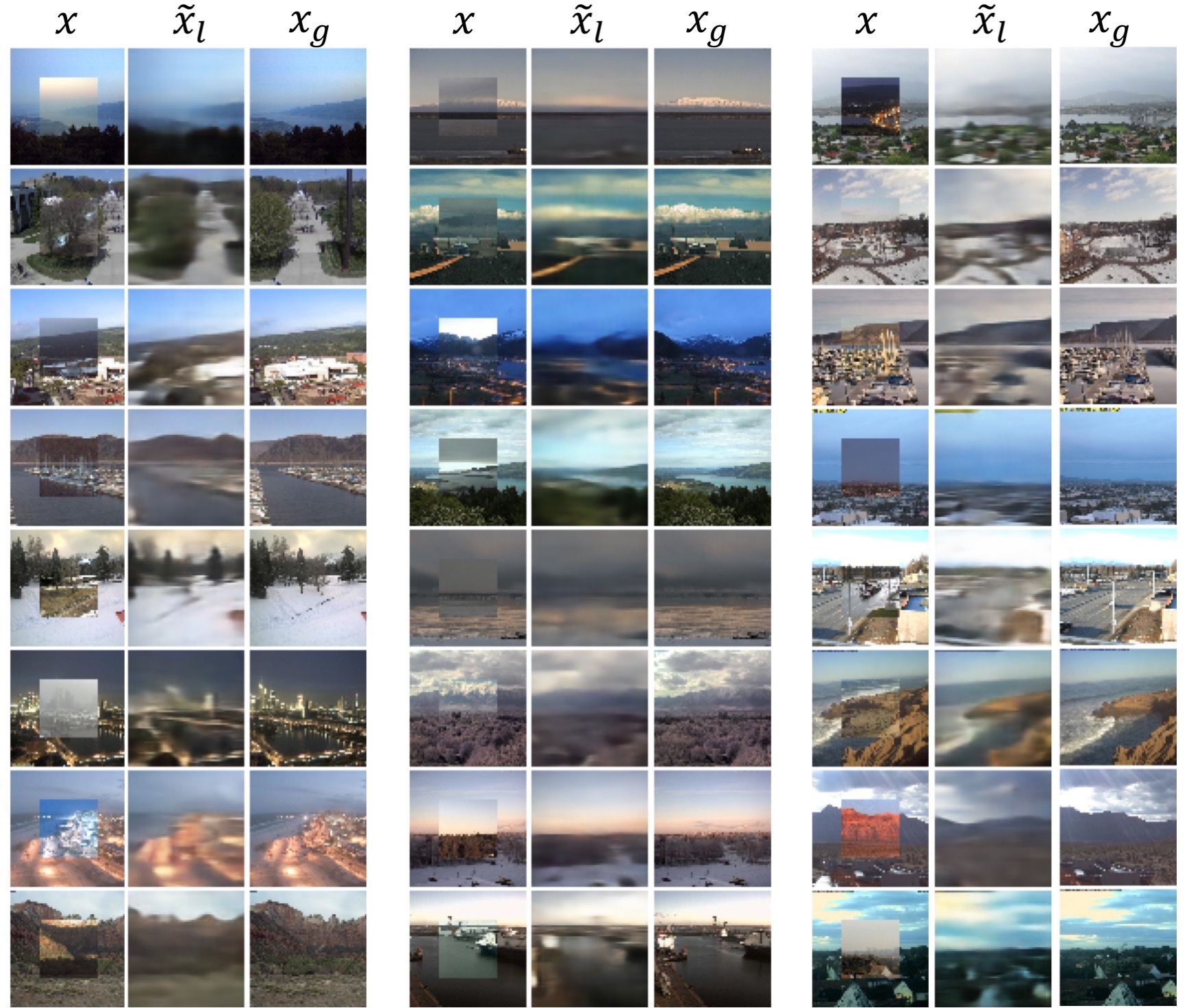}
    \caption{Image blending results generated by $G(x)$. The experiment is conducted on the Transient Attributes Database~\cite{Laffont14}. $x$ is the copy-and-paste image composited by $x_{src}$ and $x_{dst}$ with a central-squared patch as the mask. $\tilde{x}_l$ is the output of $G(x)$ with size $64 \times 64$. $x_g$ is the ground truth image used for training $G(x)$, which is the same as $x_{dst}$. Best viewed in color.}
    \label{fig:encoder_decoder_input_output_gt}
\end{figure}

\subsection{Image Blending}
\label{subsection:definition}
Given a source image $x_{src}$, a destination (target) image $x_{dst}$ and a mask image $x_{mask}$, the composite (copy-and-paste) image $x$ can be obtained by Equation~\ref{equation:copy_paste_image},
        \begin{equation} \label{equation:copy_paste_image}
            x = x_{src}\ast x_{mask} + x_{dst}\ast (1 - x_{mask}),
        \end{equation}
    where $\ast$ is element-wise multiplication operator. The goal of image blending is to generate a well-blended image $\tilde{x}$ that is semantically similar to the composite image $x$ but looks more realistic and natural with the resolution unchanged. $x$ is usually a high-resolution image.
\subsection{Framework Overview}
\label{subsection:framework}
Generating high-resolution well-blended images is hard. To tackle this problem, we propose GP-GAN, a framework for generating high-resolution and realistic images, as shown in Figure~\ref{fig:framework}. This is the first time that GANs are used for realistic high-resolution image blending to the best of our knowledge.

GP-GAN seeks a well-blended high-resolution image $\tilde{x}_h$ by optimizing a loss function consisting of a color constraint and a gradient constraint. The color constraint tries to make the generated image more realistic and natural while the gradient constraint captures the high-resolution details such as textures and edges.

The color constraint is constructed with a low-resolution realistic image $\tilde{x}_l$. To generate $\tilde{x}_l$, we propose Blending GAN $G(x)$ that learns to blend a copy-and-paste image and generate a realistic one semantically similar to the input. Once $G(x)$ is trained, we can use it to generate $\tilde{x}_l$ functioning as the color constraint.

The goal of gradient constraint is to generate the high-resolution details, including textures and edges given the composite image $x$. Their gradients directly capture textures and edges of an image. We propose Gaussian-Poisson Equation to force $\tilde{x}_h$ to have a similar gradient to $x$ while approximating the color of $\tilde{x}_l$.

GP-GAN can naturally generate realistic images in arbitrary resolution. Given a composite image $x$, we first obtain $\tilde{x}_l$ by feeding $x^1$ to $G(x)$, where $x^1$ is the coarsest scale in the Laplacian pyramid of $x$. Then we update $\tilde{x}_h^1$ by optimizing Gaussian-Poisson Equation with the closed-form solution. $\tilde{x}_h^1$ is upsampled and serves as $\tilde{x}_l$ at the finer scale in the Laplacian pyramid of $x$. The final realistic image $\tilde{x}_h$ with the same resolution as $x$ is obtained at the finest scale of the pyramid.

In Section~\ref{subsection:encoder-decoder}, we will describe the details of our Blending GAN $G(x)$. The details of GP-GAN and Gaussian-Poisson Equation will be described in Section~\ref{subsection:poisson_gan}.

\subsection{Blending GAN}
\label{subsection:encoder-decoder}

We seek a low-resolution well-blended image $\tilde{x}_l$ that is visually realistic and semantically similar to the input image. A straightforward way is to train a conditional GAN and use the generator to produce realistic images. Since we have both the input image and the corresponding ground truth $x_g$, we aim to train a generator in a supervised way. To achieve this goal, we propose Blending GAN $G(x)$, which leverages the unsupervised Wasserstein GAN~\cite{arjovsky2017wasserstein} for supervised learning tasks. The proposed Blending GAN is different from Wasserstein GAN in that it has a proper constructed auxiliary loss and dedicated designed architecture.

Recent works discuss various loss functions for image processing tasks, for instance, $l_1$ loss~\cite{zhao2015l2}, $l_2$ loss, and perceptual loss~\cite{johnson2016perceptual}. $l_1$ and $l_2$ loss can accelerate the training process but tend to produce blurry images. The perceptual loss is good at generating high-quality images but is time and memory consuming. We employ $l_2$ loss in this paper because it could accelerate the training process and generate sharp and realistic images when combined with GANs~\cite{Isola/cvpr2017}. The combined loss function is defined as follows:
        \begin{equation} \label{equation:regression_loss}
            L(x, x_g) = \lambda L_{l_2}(x, x_g) + (1 - \lambda) L_{adv}(x, x_g),
        \end{equation}
    where $\lambda$ is $0.999$ in our experiment. $L_{l_2}$ is defined as follows:
        \begin{equation} \label{equation:regression_l2_loss}
            L_{l_2}(x, x_g) = \|G(x) - x_g\|_2^2,
        \end{equation}
    and $L_{adv}$ is defined as follows:
        \begin{equation} \label{equation:regression_gan_loss}
            L_{adv}(x, x_g) = \max_D{E_{x\in\chi}[D(x_g) - D(G(x))]}.
        \end{equation}

The architecture of Blending GAN $G(x)$ is shown in Figure~\ref{fig:encoder-decoder}, which is motivated by~\cite{pathak2016context}. We find that a network with only convolutional layers could not learn to blend composite images for the lack of global information across the whole image. Thus we replace the channel-wise fully connected layer used in~\cite{pathak2016context} with standard fully connected layers.

The architecture of the discriminator $D(x)$ is shown in Figure~\ref{fig:discriminator}. We apply the batch normalization~\cite{ioffe2015batch} and leaky ReLU after each convolution except for the first layer and the last layer. The first layer employs convolution and leaky ReLU, while the last layer contains convolution only.

Training such a network needs massive data. The copy-and-paste images are easy to collect, but the ground truth images $x_g$ could only be obtained by expert users with image editing software, which is time-consuming. Alternatively, we use $x_{dst}$ to approximate $x_g$, since $x_{src}$ and $x_{dst}$ in our experiment are photos of the same scene under different conditions, \eg season, weather, time of day, see Section~\ref{ph:datasets} for details. Through this way, we obtain massive composite images and the corresponding ground truth, as shown in Figure~\ref{fig:encoder_decoder_input_output_gt}.
\subsection{Gaussian-Poisson Equation}
\label{subsection:poisson_gan}

The proposed Blending GAN can only generate low-resolution images, as shown in Figure~\ref{fig:encoder_decoder_input_output_gt}. Even for slightly larger images, the results tend to be blurry with unpleasant artifacts, which is unsuitable for image blending task. Since the task usually needs to combine several high-resolution images and blend them into one realistic image with the resolution unchanged. To make use of the realistic images generated by Blending GAN, we propose Gaussian-Poisson Equation fashioned by the well-known Laplacian pyramid~\cite{burt1983laplacian} for generating high-resolution and realistic images.

We observe that although our Blending GAN cannot produce high-resolution images, the generated image $\tilde{x}_l$ is natural and realistic as a low-resolution image. So we can seek a high-resolution and realistic image $\tilde{x}_h$ by approximating the color of $\tilde{x}_l$ while capturing rich details like textures and edges in the original high-resolution image $x$. Such requirements are formulated into two constraints: one is the color constraint, while the other is the gradient constraint. The color constraint forces $\tilde{x}_h$ to have a similar color to $\tilde{x}_l$, which can be achieved by generating an image with the same low-frequency signals as $\tilde{x}_l$. The simplest way to extract the low-frequency signals is using a Gaussian filter. The gradient constraint tries to restore the high-resolution details, which is the same as forcing $\tilde{x}_h$ and $x$ to have the same high-frequency signals. This step could be implemented by using the divergence operator.

Formally, we need to optimize the objective function defined as follows:
        \begin{equation} \label{equation:objective_func}
            H(x_h) = P(x_h) + \beta C(x_h).
        \end{equation}
    $P(x_h)$ is inspired by the well-known Poisson Equation~\cite{perez2003poisson} and is defined as follows:
        \begin{equation} \label{equation:poisson}
            P(x_h) = \int_T\|\textbf{div}\,v-\Delta x_h\|_2^2\,dt,
        \end{equation}
    $C(x_h)$ is defined as follows:
        \begin{equation} \label{equation:color}
            C(x_h) = \int_T\|g(x_h)-\tilde{x}_l\|_2^2\,dt,
        \end{equation}
    and $\beta$ represents the color preserving parameter. We set $\beta$ to $1$ in our experiment.
    In Equation~\ref{equation:poisson}, $T$ represents the whole image region, $\textbf{div}$ represents the divergence operator and $\Delta$ represents the Laplacian operator. $v$ is defined as follows:
        \begin{equation} \label{equation:v}
            v^{i} =
                \begin{cases}
                       \nabla x_{src}^i & \text{if } x_{mask}^{i} = 1 \\
                       \nabla x_{dst}^i & \text{if } x_{mask}^{i} = 0
                \end{cases},
        \end{equation}
    where $\nabla$ is the gradient operator. Gaussian filter is used in Equation~\ref{equation:color} and is denoted as $g(x_h)$.
    The discretized version of Equation~\ref{equation:objective_func} is defined as follows:
        \begin{equation} \label{equation:objective_dis}
            H(x_h) = \|u - LX_h\|_2^2 + \lambda\|GX_h-\tilde{X}_l\|_2^2,
        \end{equation}
    where $u$ is the discretized divergence of $v$, $L$ is the matrix of the Laplacian operator, and $G$ represents the Gaussian filter. $X_h$ and $\tilde{X}_l$ are the vector representation of $x_h$ and $\tilde{x}_l$. The closed-form solution for minimizing the cost function of Equation~\ref{equation:objective_dis} could be obtained in the same manner as~\cite{frankot1988method}.

We integrate the closed-form solution for optimizing Equation~\ref{equation:objective_dis} and the Laplacian pyramid into our final high-resolution image blending algorithm, which is described by Algorithm~\ref{algo:high_res}. Given a high-resolution input image $x_{src}$, $x_{dst}$ and $x_{mask}$, we first generate the low-resolution realistic image $\tilde{x}_l$ using Blending GAN $G(x)$. Then we generate Laplacian pyramids $x_{src}^s, x_{dst}^s, x_{mask}^s, s=1,2,...,S$, where $S$ is the number of scales. $s = 1$ is the coarsest scale and $s = S$ is the original resolution. We update $x_h^s$ by optimizing Equation~\ref{equation:objective_dis} at each scale and set $\tilde{x}_l$ to be upsampled $x_h^s$. The final realistic image $\tilde{x}_h$ with the unchanged resolution is set to be $x_h^S$.
\begin{algorithm}
\caption{High-Resolution Image Blending Framework GP-GAN}
\label{algo:high_res}
\SetKwInOut{Input}{Input}
\SetKwInOut{Output}{Output}
\Input{Source image $x_{src}$, destination image $x_{dst}$, mask image $x_{mask}$ and trained Blending GAN $G(x)$}

Compute Laplacian Pyramid for $x_{src}$, $x_{dst}$ and $x_{mask}$

Compute $\tilde{x}_{l}$ using $G(x)$

\For{$s \in [1,2,...,S]$}{
    Updating $x_h^s$ by optimizing Equation~\ref{equation:objective_dis} using the closed form solution given $x_{src}^s$, $x_{dst}^s$, $x_{mask}^s$ and $\tilde{x}_{l}$

    Set $\tilde{x}_{l}$ to be upsampled $x_h^{s}$
}

Return $x_h^S$
\end{algorithm}
\section{Experiments}
\label{section:experiment}
In this section, the datasets for the experiments are introduced firstly. Then the training configurations and experimental settings are described. Finally, the effectiveness of our method are shown quantitatively and visually by comparing with other methods.

\subsection{Dataset}
\label{ph:datasets}
\begin{figure}[h]
	\centering
    \begin{subfigure}{0.48\linewidth}
        \includegraphics[width=\linewidth]{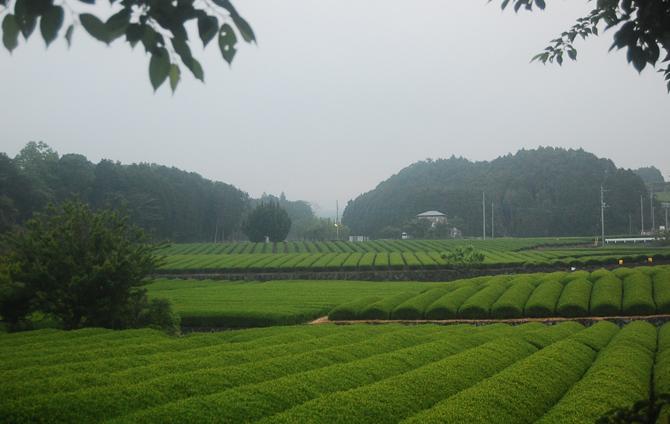}
        \caption{Source Image} \label{fig:transient_attributes_database_a}
    \end{subfigure}
    \hspace*{\fill}
    \begin{subfigure}{0.48\linewidth}
        \includegraphics[width=\linewidth]{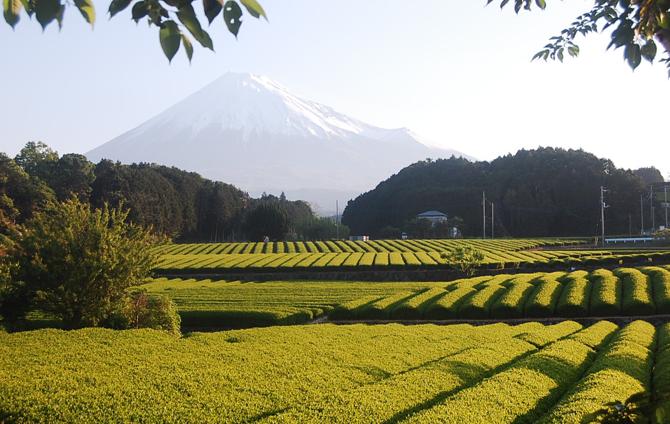}
        \caption{Target Image} \label{fig:transient_attributes_database_b}
    \end{subfigure}
    \\
    \begin{subfigure}{0.48\linewidth}
        \includegraphics[width=\linewidth]{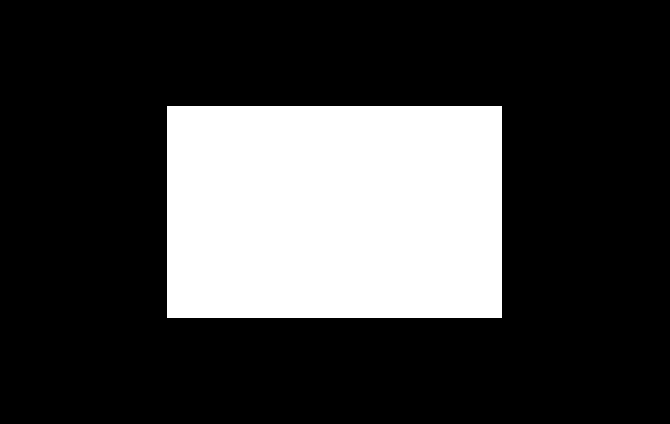}
        \caption{Mask A} \label{fig:transient_attributes_database_c}
    \end{subfigure}
    \hspace*{\fill}
    \begin{subfigure}{0.48\linewidth}
        \includegraphics[width=\linewidth]{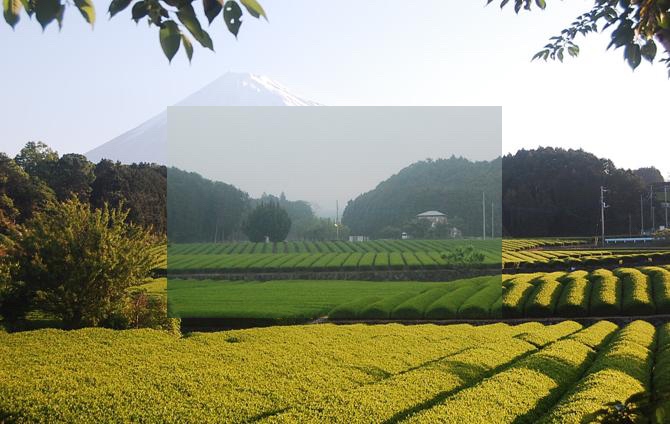}
        \caption{Composite Image A} \label{fig:transient_attributes_database_d}
    \end{subfigure}
    \\
    \begin{subfigure}{0.48\linewidth}
        \includegraphics[width=\linewidth]{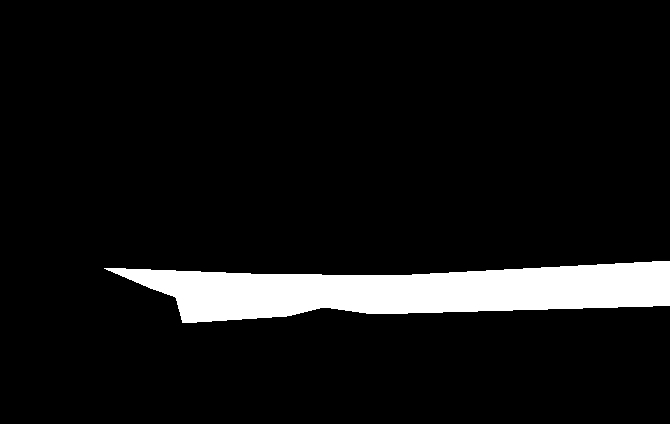}
        \caption{Mask B} \label{fig:transient_attributes_database_e}
    \end{subfigure}
    \hspace*{\fill}
    \begin{subfigure}{0.48\linewidth}
        \includegraphics[width=\linewidth]{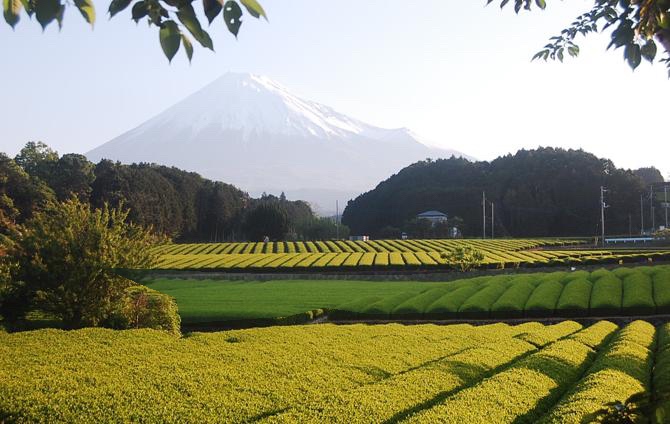}
        \caption{Composite Image B} \label{fig:transient_attributes_database_f}
    \end{subfigure}
    \caption{Transient Attributes Database. (a) $x_{src}$ and (b) $x_{dst}$ are from the same webcam but in different seasons. (c) is the central-squared mask and (d) is the corresponding composite image. (e) is the object-level mask annotated with LabelMe and (f) is the corresponding composite image. Best viewed in color.}
\label{fig:transient_attributes_database}
\end{figure}

Transient Attributes Database~\cite{Laffont14} contains 8,571 images from 101 webcams. In each webcam, there are well-aligned 60-120 images with severe appearance changes caused by weather, time of day, and season, as shown in Figure~\ref{fig:transient_attributes_database_a} and Figure ~\ref{fig:transient_attributes_database_b}. 

For training $G(x)$, we randomly select 2 images from the same camera as $x_{src}$ (Figure~\ref{fig:transient_attributes_database_a}) and $x_{dst}$ (Figure ~\ref{fig:transient_attributes_database_b}). As for the ground truth $x_g$, we use $x_{dst}$ to approximate it since images under the same webcam is perfect-aligned. $x_{mask}$ is a binary image with a central-squared patch, as shown in Figure~\ref{fig:transient_attributes_database_c}. The composite copy-and-paste image is then obtained by Equation~\ref{equation:copy_paste_image}, as shown in Figure~\ref{fig:transient_attributes_database_d}. Although $G(x)$ is trained with the central-squared patch as the mask, our experiments show that it is still able to generate well-blended images for inputs with arbitrary masks. 

To evaluate our method with arbitrary masks, we first manually annotate object-level masks for Transient Attributes Database with the LabelMe~\cite{russell2008labelme} annotation tool. Then we use the object-level masks to composite the copy-and-paste images, which are used to evaluate different image blending methods. The annotated mask and corresponding composite image are shown in Figure~\ref{fig:transient_attributes_database_e} and Figure~\ref{fig:transient_attributes_database_f}.

\subsection{Implementation Details}
Our method is implemented with Chainer~\cite{tokui2015chainer}. To train Blending GAN, we employ ADAM~\cite{kingma2014adam} for optimization, where $\alpha$ is set to $0.002$, and $\beta_1$ is set to 0.5. We randomly generate 150K images from Transient Attributes Database using the central-squared patch as the mask. Then the network is trained for 25 epochs with batch size 64.

\subsection{Quantitative Comparisons}
Our method is compared with three classic image blending approaches. Poisson Image Editing (PB)~\cite{perez2003poisson} and its improved version Modified Poisson Image Editing (MPB)~\cite{tanaka2012seamless} are selected as baselines because both of them employ Poisson Equation in their solutions as our method does. We also compare with multi-splines blending (MSB)~\cite{szeliski2011fast} for its effectiveness and extensive usage.

We first show the quantitative results of our method with realism score as the metric. Realism score is produced by RealismCNN~\cite{zhu2015learning}, which predicts the visual realism of an image regarding color, lighting, and texture compatibility.

Our method is evaluated on 500 images that are randomly sampled from Transient Attributes Database with the annotated masks. The average realism scores for our method and the baselines are shown in Table~\ref{table:realism_score}, where our method outperforms all the baselines. We attribute this to the nature of our method because it can learn what contributes to a realistic and natural image through adversarial learning on large datasets. The average scores are negative for all evaluated methods, which shows that many blended images are still not realistic. This suggests that there are still many improvements to be made for image blending algorithms.

\begin{figure}[h]
	\centering
    \includegraphics[width=\linewidth]{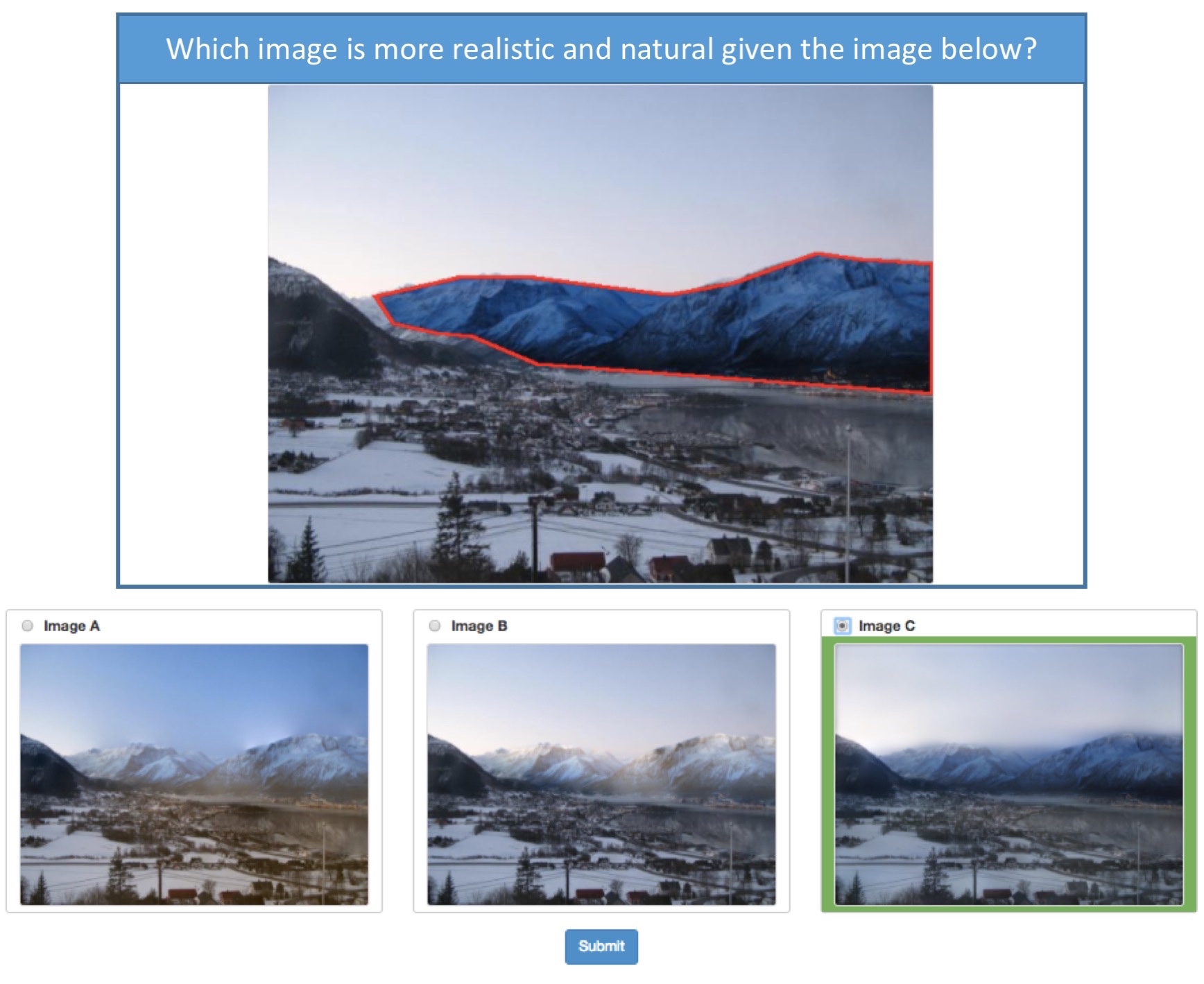}
    \caption{The user interface for user study on Amazon Mechanical Turk. Followed by the composite image with $x_{src}$ circled out, three blended results generated by different algorithms are shown to subjects, and the most realistic one is picked out. Best viewed in color.}
\label{fig:user_study}
\end{figure}

\begin{figure}[h]
	\centering
    \begin{subfigure}{0.32\linewidth}
        \includegraphics[width=\linewidth]{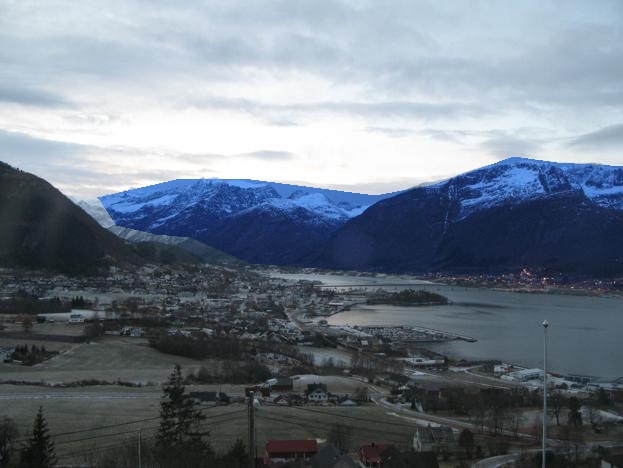}
        \caption{Input} \label{fig:role_low_res_a}
    \end{subfigure}
    \hspace*{\fill}
    \begin{subfigure}{0.32\linewidth}
        \includegraphics[width=\linewidth]{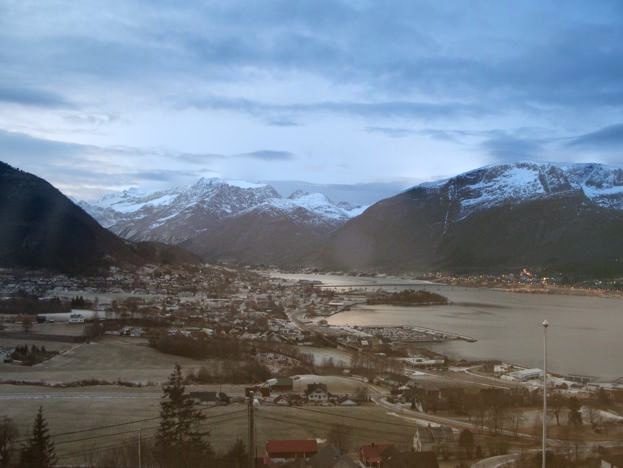}
        \caption{$x^1$} \label{fig:role_low_res_b}
    \end{subfigure}
    \hspace*{\fill}
    \begin{subfigure}{0.32\linewidth}
        \includegraphics[width=\linewidth]{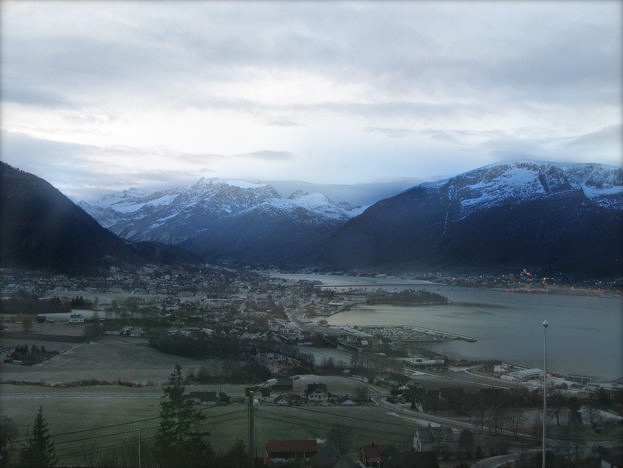}
        \caption{$\tilde{x}_l$} \label{fig:role_low_res_c}
    \end{subfigure}
    \caption{Role of Blending GAN. (a) is a copy-and-paste image. (b) employs the down-sampled $x^1$ as the color constraint. (c) uses the output of Blending GAN $\tilde{x}_l$ as the color constraint. Best viewed in color.}
\label{fig:role_low_res}
\end{figure}

\begin{figure*}[h]
	\centering
    \includegraphics[height=0.14\linewidth]{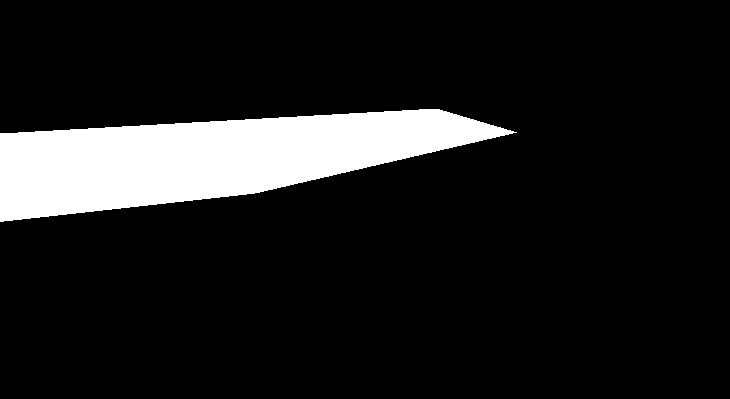}
    \includegraphics[height=0.14\linewidth]{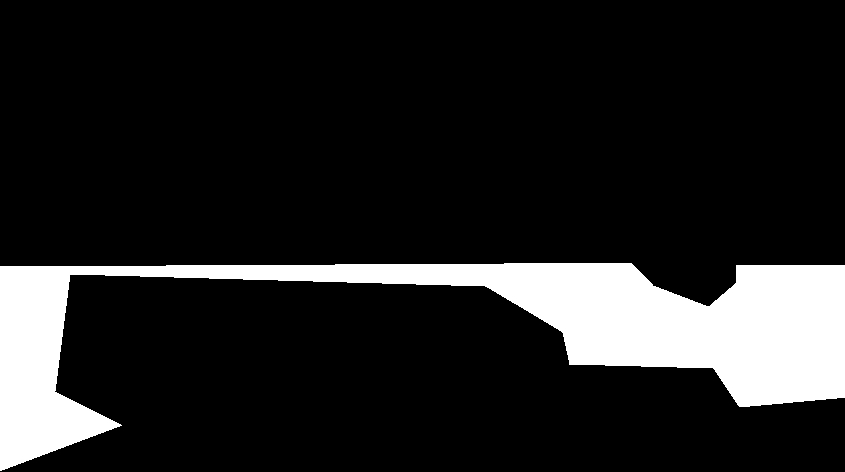}
    \includegraphics[height=0.14\linewidth]{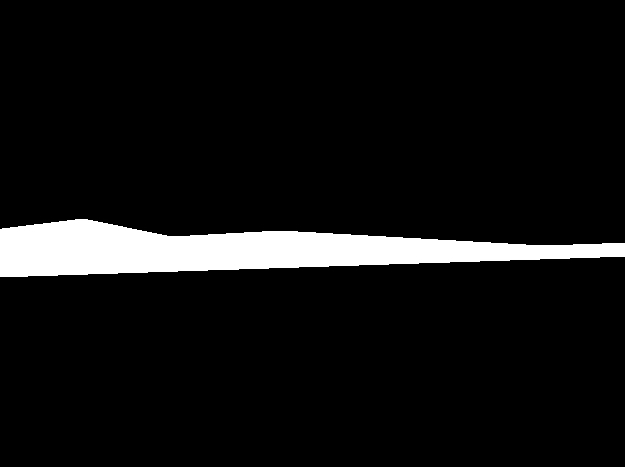}
    \includegraphics[height=0.14\linewidth]{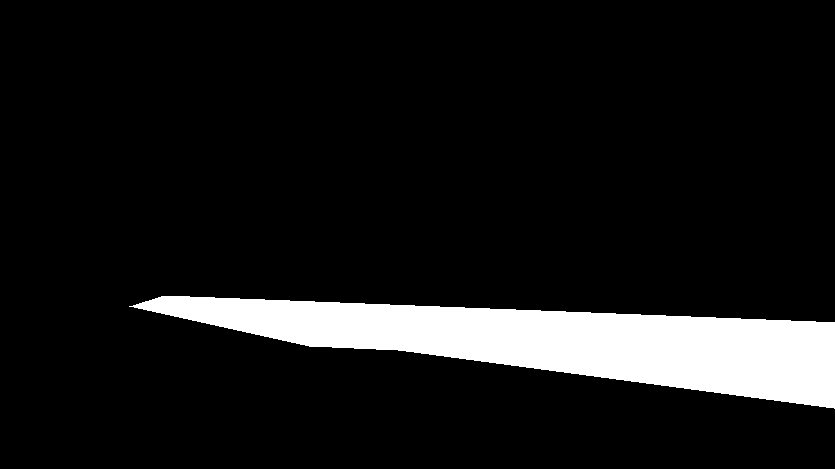}
    \\
    \includegraphics[height=0.14\linewidth]{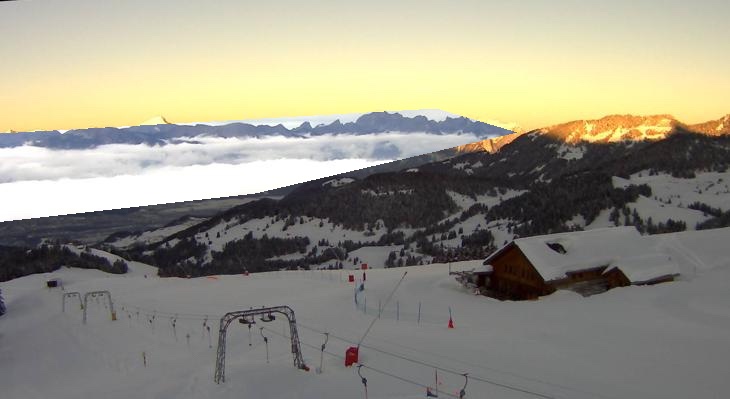}
    \includegraphics[height=0.14\linewidth]{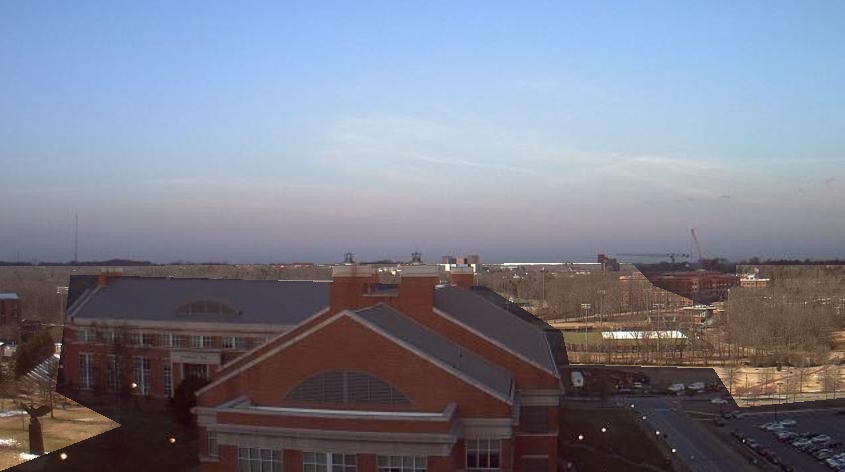}
    \includegraphics[height=0.14\linewidth]{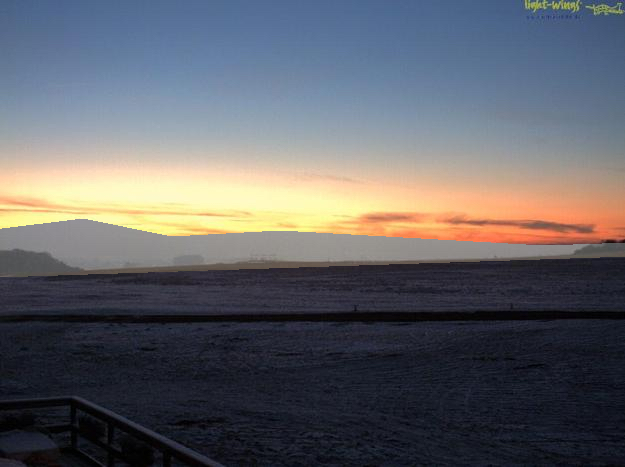}
    \includegraphics[height=0.14\linewidth]{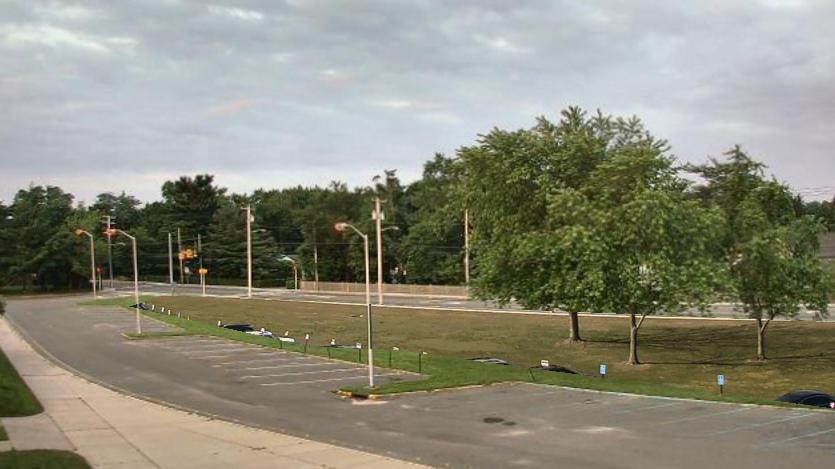}
    \\
    \includegraphics[height=0.14\linewidth]{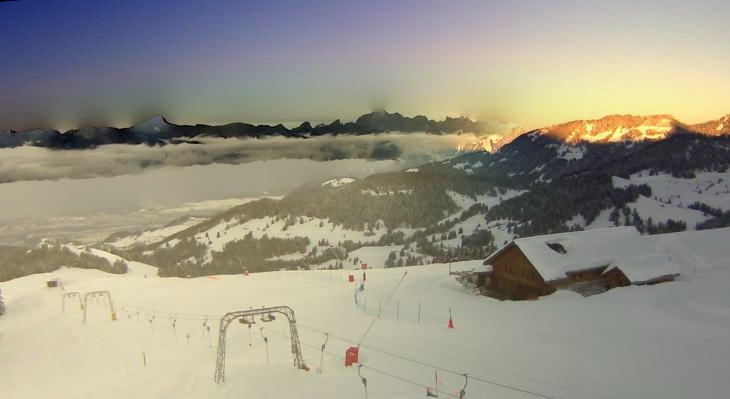}
    \includegraphics[height=0.14\linewidth]{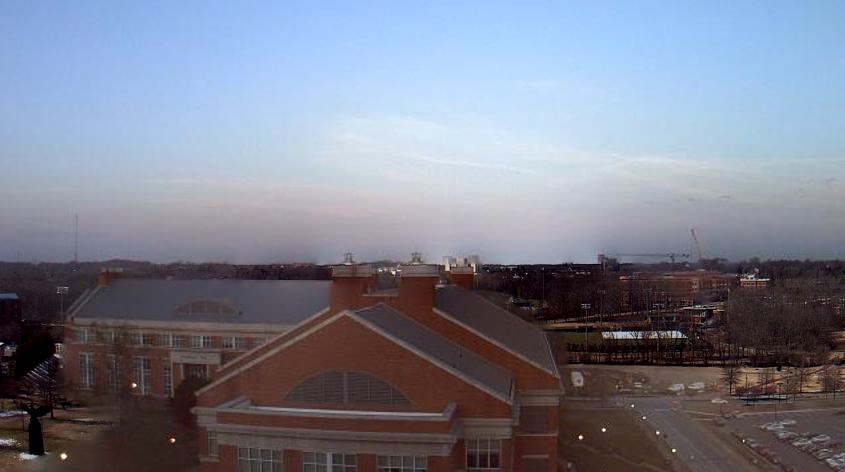}
    \includegraphics[height=0.14\linewidth]{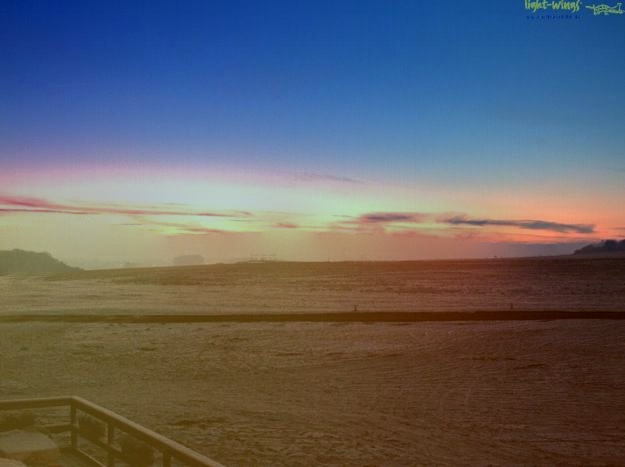}
    \includegraphics[height=0.14\linewidth]{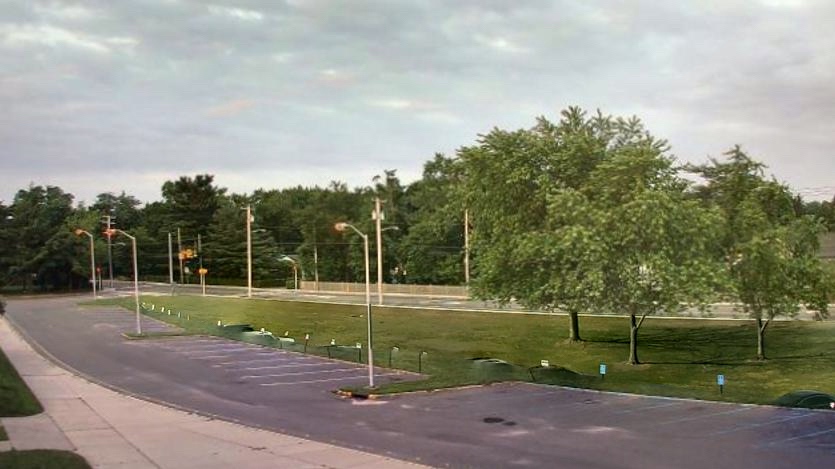}
    \\
    \includegraphics[height=0.14\linewidth]{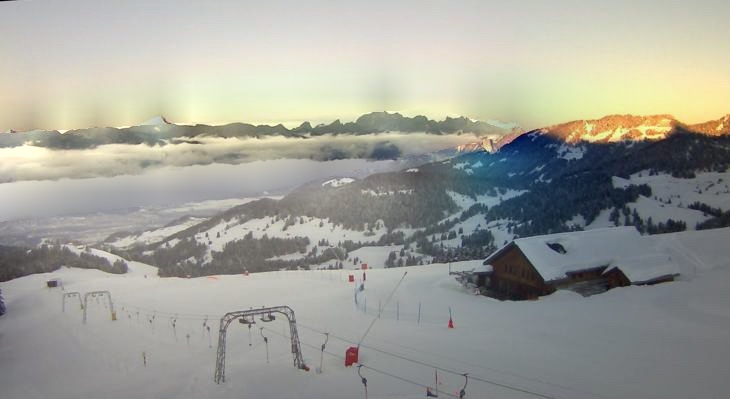}
    \includegraphics[height=0.14\linewidth]{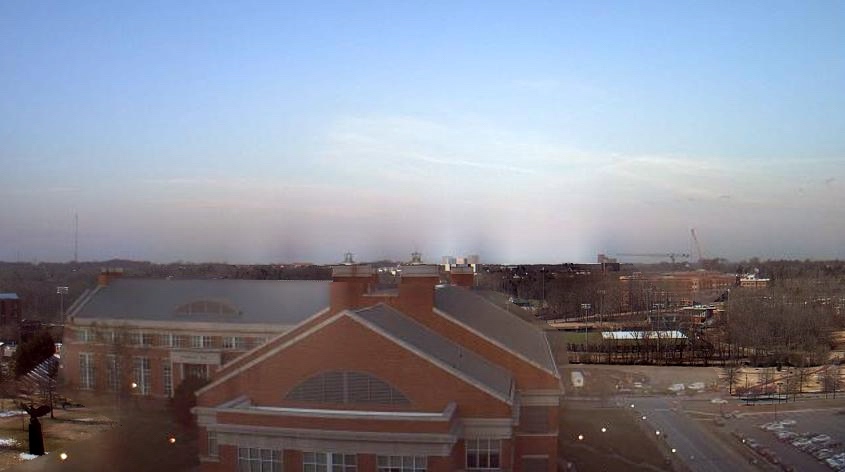}
    \includegraphics[height=0.14\linewidth]{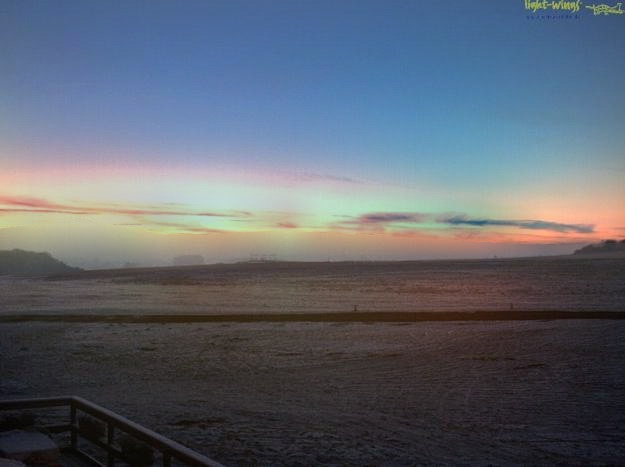}
    \includegraphics[height=0.14\linewidth]{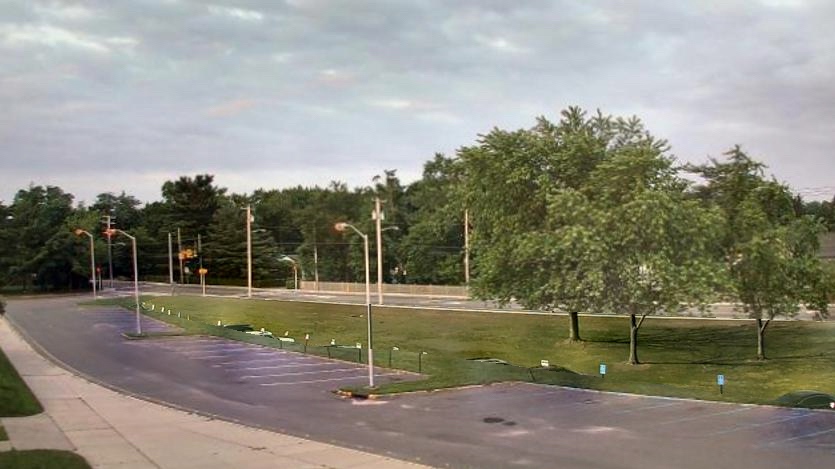}
    \\
    \includegraphics[height=0.14\linewidth]{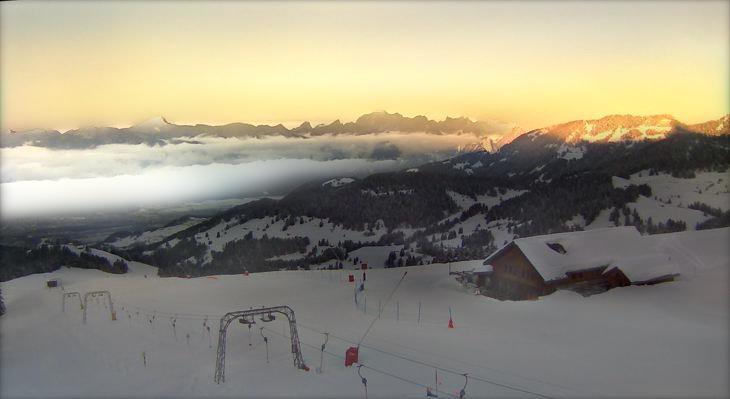}
    \includegraphics[height=0.14\linewidth]{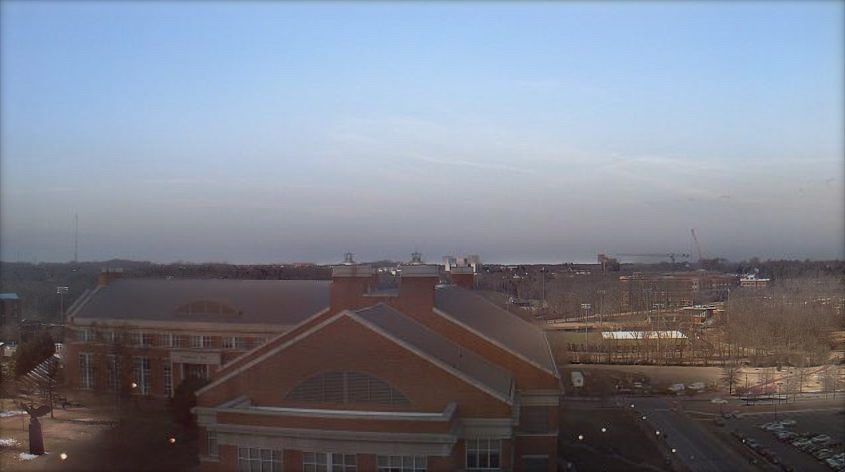}
    \includegraphics[height=0.14\linewidth]{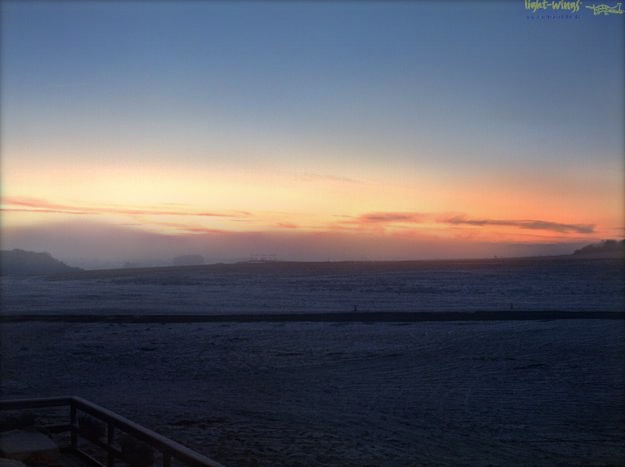}
    \includegraphics[height=0.14\linewidth]{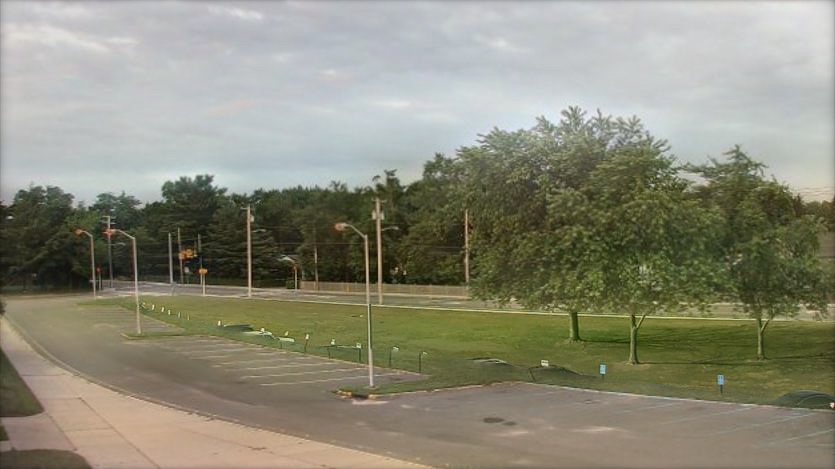}
    \caption{Results of our high-resolution blending algorithm compared with baseline methods. From top to bottom: annotated object-level mask, composite copy-and-paste image, MPB, MSB, and GP-GAN(ours). Results of baseline methods have severe bleedings, illumination inconsistencies, or other artifacts, while GP-GAN produces pleasant, realistic images. Best viewed in color.}
\label{fig:visually_comparision}
\end{figure*}

\begin{figure*}[h]
	\centering
    \includegraphics[height=0.33\linewidth]{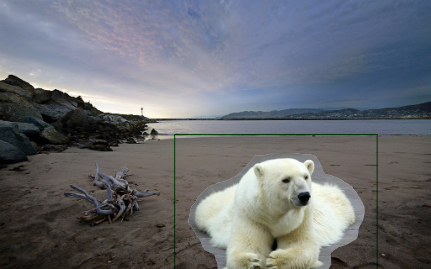}
    \includegraphics[height=0.33\linewidth]{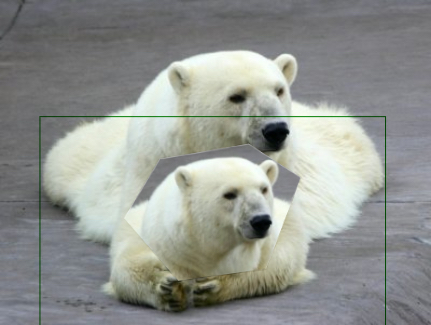}\\
    \includegraphics[height=0.33\linewidth]{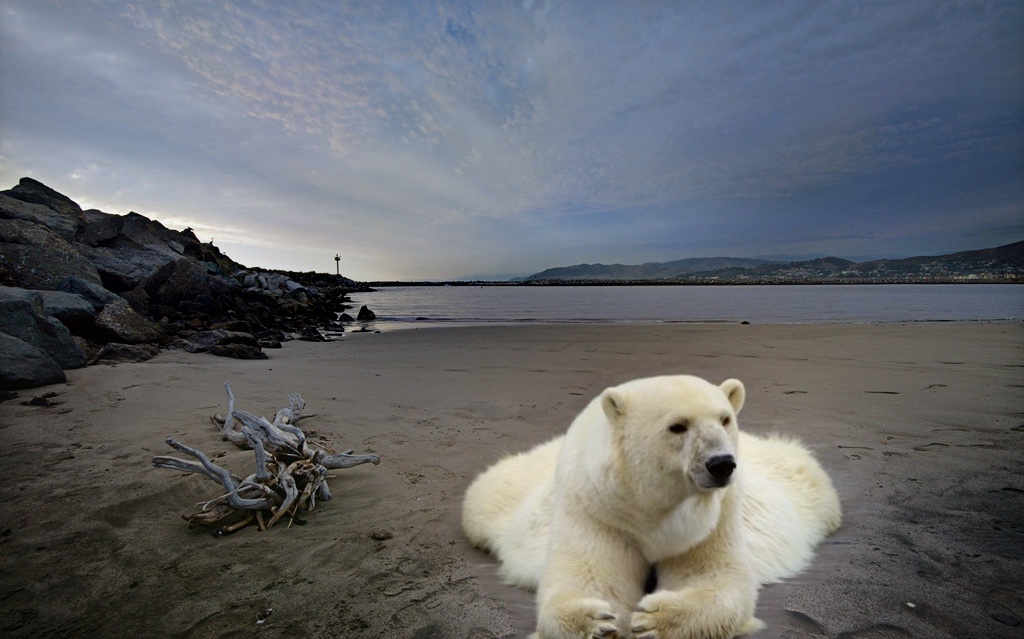}
    \includegraphics[height=0.33\linewidth]{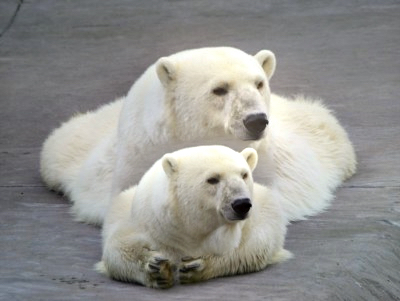}
    \caption{Results of GP-GAN on real images. The top is the copy-and-paste images and the bottom is the blended images. Best viewed in color.}
\label{fig:real}
\end{figure*}

\subsection{User Study} Realism scores show the effectiveness of our method. Since image blending is a user-oriented task, it is essential to conduct user study for evaluation. We employ Amazon Mechanical Turk to collect user assessments. Each time, a composite image $x$ is shown to the subjects followed by three blended images produced by three different algorithms. The subjects are told to pick the most realistic image among these three blended images, as shown in Figure~\ref{fig:user_study}. The statistical result of user study is reported in Table~\ref{table:user_study}. GP-GAN is preferred by the majority of users, which is consistent with the result of realism scores in Table~\ref{table:realism_score}.

\begin{table}
\caption{Realism scores for our method and the baselines (higher is better). GP-GAN outperforms all the baselines.}
\label{table:realism_score}
\begin{tabular}{lccccc}
\toprule
Method & Input & PB\cite{perez2003poisson} & MPB\cite{tanaka2012seamless} & MSB\cite{szeliski2011fast} & Ours \\
\midrule
Score & -0.696 & -0.192 & -0.151 & -0.140 & -0.069\\
\bottomrule
\end{tabular}
\end{table}

\begin{table}
\caption{User study result. 4 image blending algorithms are compared on Amazon Mechanical Turk. Our method GP-GAN obtains most votes, which is consistent with the result of the realism scores.}
\label{table:user_study}
\begin{tabular}{lccc}
\toprule
Method & Total votes & Average votes & Std.\\
\midrule
PB\cite{perez2003poisson} & 527 & 1.054 & 1.065\\
MPB\cite{tanaka2012seamless} & 735 & 1.470 & 1.173\\
MSB\cite{szeliski2011fast} & 770 & 1.540 & 1.271\\
\midrule
GP-GAN & 947 & 1.894 & 1.311\\
\bottomrule
\end{tabular}
\end{table}

\subsection{Role of Blending GAN} The output of Blending GAN serves as the color constraint. In this section, we demonstrate the role of Blending GAN by replacing $\tilde{x}_l$ with the down-sampled composite image $x^1$. The blended results with either $\tilde{x}_l$ or $x^1$ as the color constraint are compared. As shown in Figure~\ref{fig:role_low_res}, the blended image tends to have more bleedings and illumination inconsistencies if $\tilde{x}_l$ is replaced by $x^1$, which shows the usefulness of low-resolution natural images in our method.

\subsection{Qualitative Comparisons} Finally, we demonstrate the results of our high-resolution image blending algorithm visually by comparing with MPB and MSB. As shown in Figure~\ref{fig:visually_comparision}, our method tends to generate realistic results while preserving the appearance of both $x_{src}$ and $x_{dst}$. Compared to the baseline methods, there are nearly no bleedings or illumination inconsistencies in our results while all the baseline methods have more or fewer bleedings and artifacts.

Our method can also be applied to real images in high resolution, as shown in Figure~\ref{fig:real}.
\section{Conclusion}

We advanced the state-of-the-art in image blending by combining the ideas from the generative model GANs and gradient-based approaches. Our insight is, on the one hand, GANs are good at generating natural images from a particular distribution but weak in capturing the high-frequency image details like textures and edges. On the other hand, the gradient-based methods perform well at generating high-resolution images with local consistency, although the generated images tend to be unnatural and have many artifacts. GANs and gradient-based methods should be integrated. Hence, this integration would result in an image blending system that overcomes the drawbacks of both approaches. Our system can also be useful for image-to-image translation task. Despite the effectiveness, our algorithm fails to generate realistic images when the composite images are far away from the distribution of the training dataset. We aim to address this issue in future work.
%
\begin{acks}
This work is funded by the National Natural Science Foundation of China (Grand No. 61876181, 61721004, 61403383) and the Projects of Chinese Academy of Sciences (Grand QYZDB-SSW-JSC006 and Grand 173211KYSB20160008).
\end{acks}

%
\bibliographystyle{ACM-Reference-Format}
\balance
\bibliography{gp_gan}

\onecolumn
\section*{A. Visual Results}
\begin{figure*}[hb]
\begin{center}
    \includegraphics[height=0.15\linewidth]{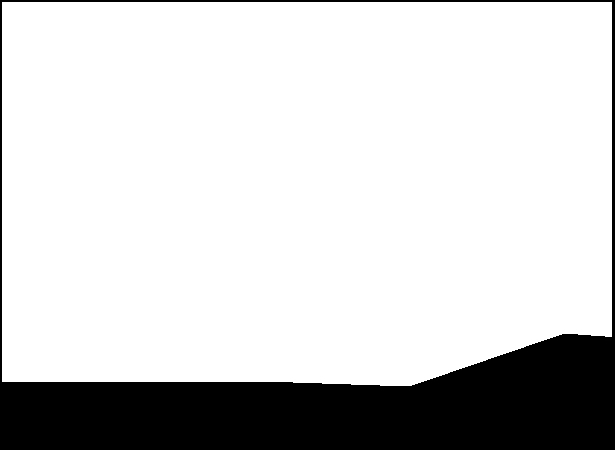}
    \includegraphics[height=0.15\linewidth]{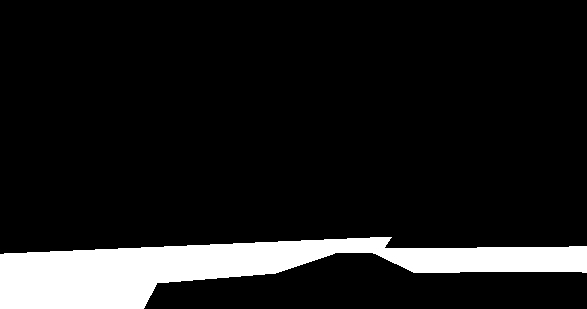}
    \includegraphics[height=0.15\linewidth]{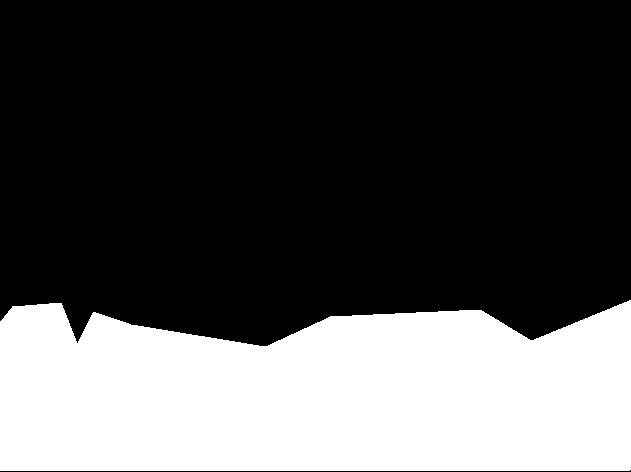}
    \includegraphics[height=0.15\linewidth]{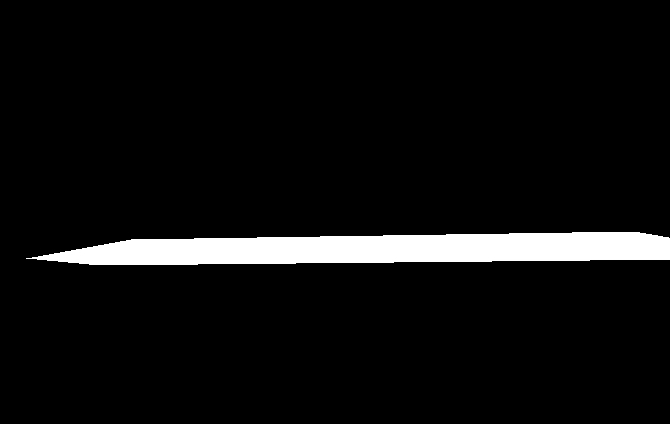}
    \\
    \includegraphics[height=0.15\linewidth]{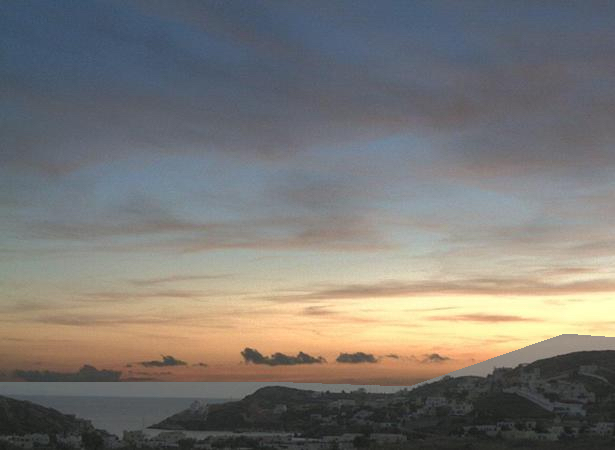}
    \includegraphics[height=0.15\linewidth]{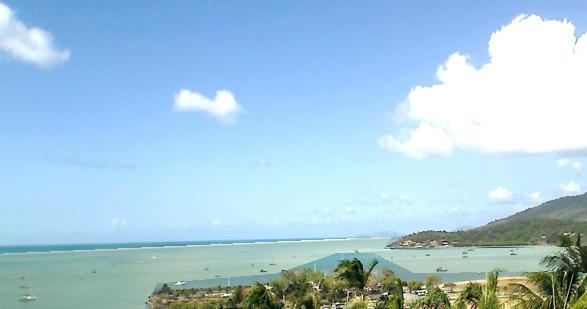}
    \includegraphics[height=0.15\linewidth]{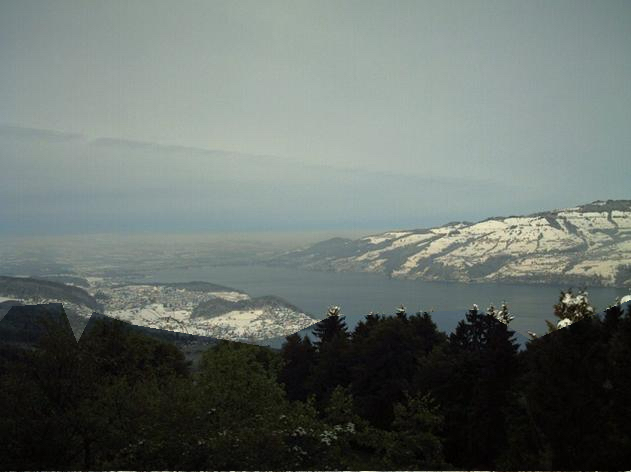}
    \includegraphics[height=0.15\linewidth]{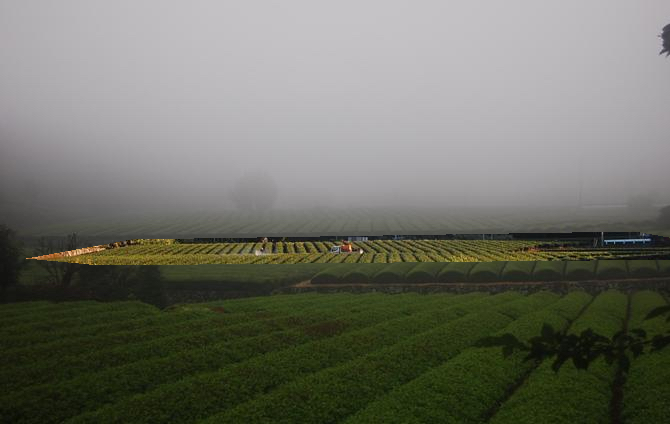}
    \\
    \includegraphics[height=0.15\linewidth]{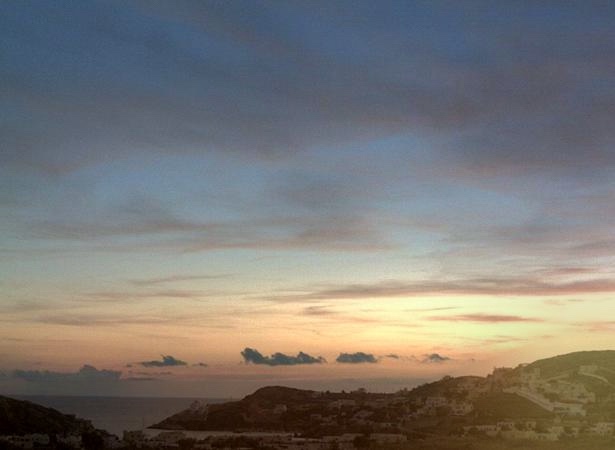}
    \includegraphics[height=0.15\linewidth]{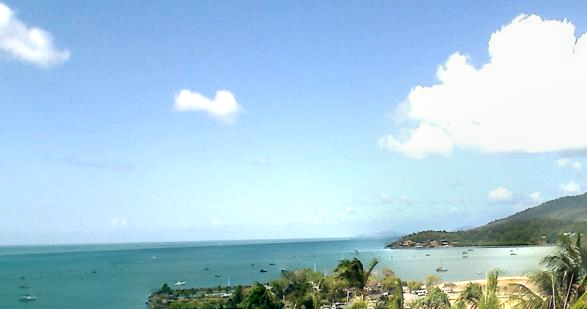}
    \includegraphics[height=0.15\linewidth]{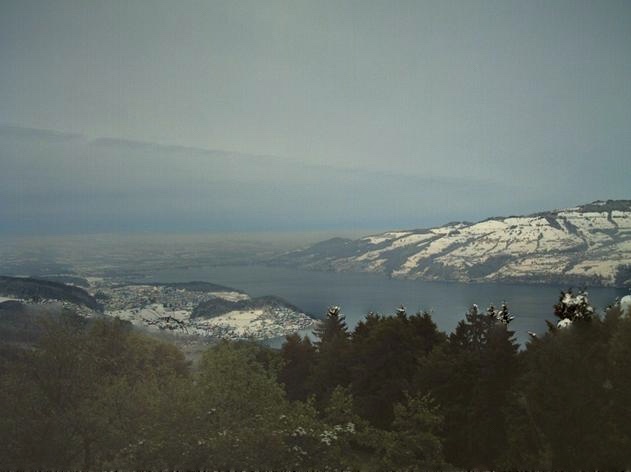}
    \includegraphics[height=0.15\linewidth]{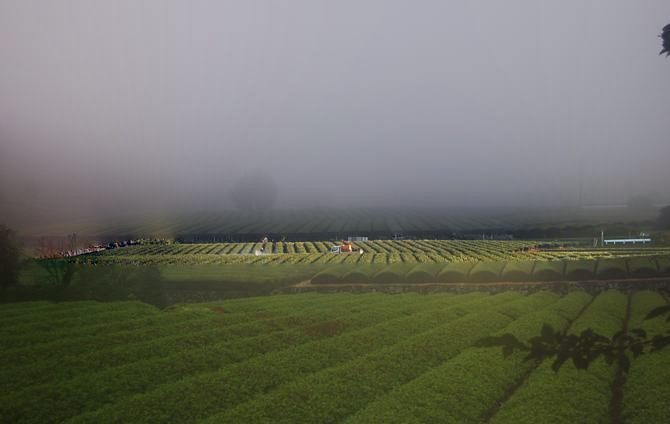}
    \\
    \includegraphics[height=0.15\linewidth]{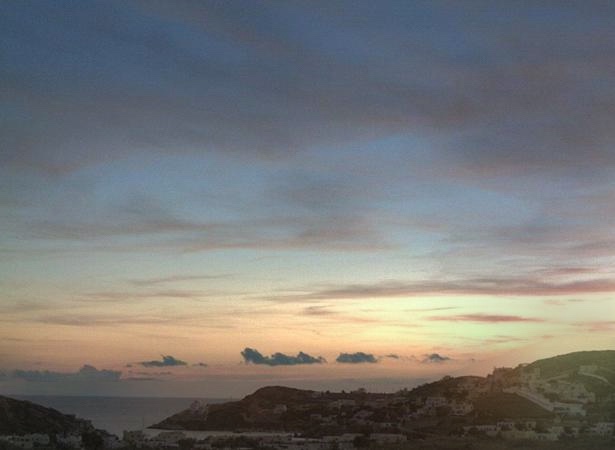}
    \includegraphics[height=0.15\linewidth]{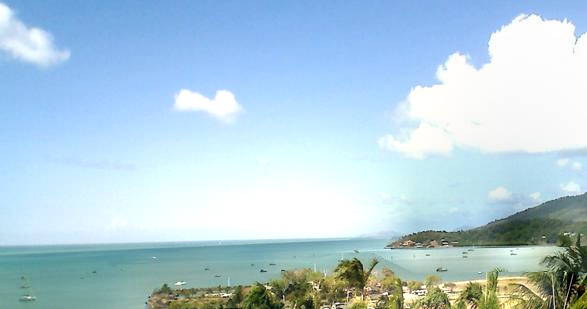}
    \includegraphics[height=0.15\linewidth]{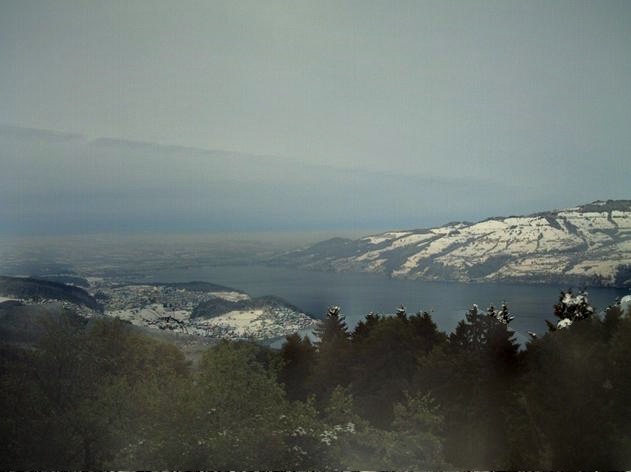}
    \includegraphics[height=0.15\linewidth]{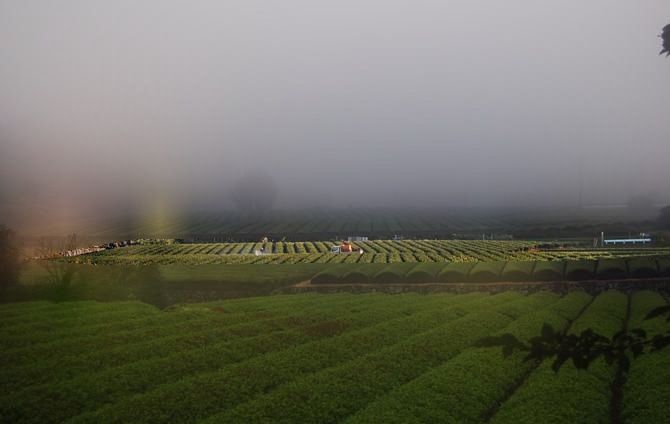}
    \\
    \includegraphics[height=0.15\linewidth]{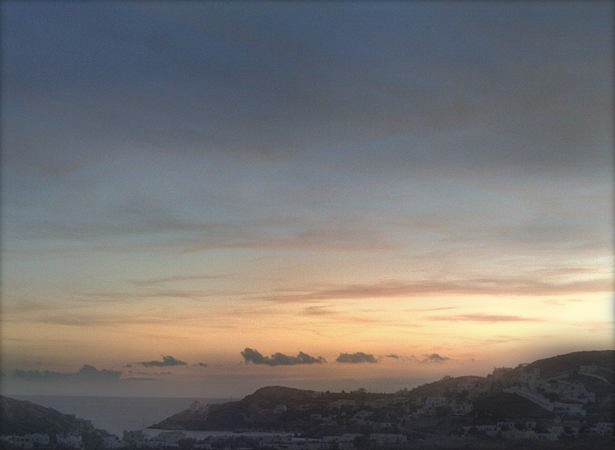}
    \includegraphics[height=0.15\linewidth]{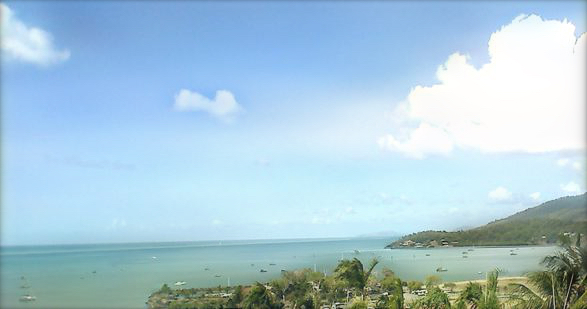}
    \includegraphics[height=0.15\linewidth]{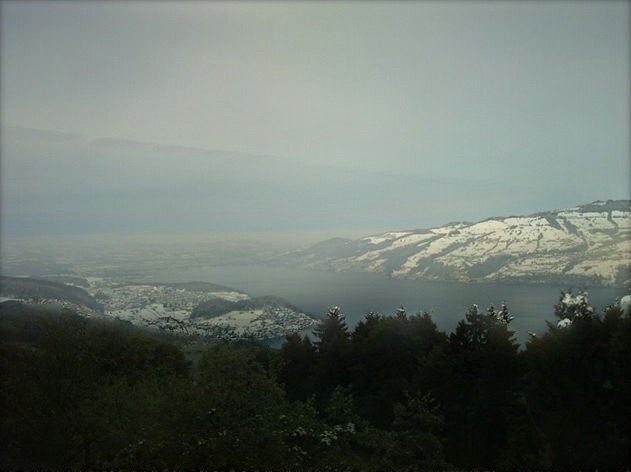}
    \includegraphics[height=0.15\linewidth]{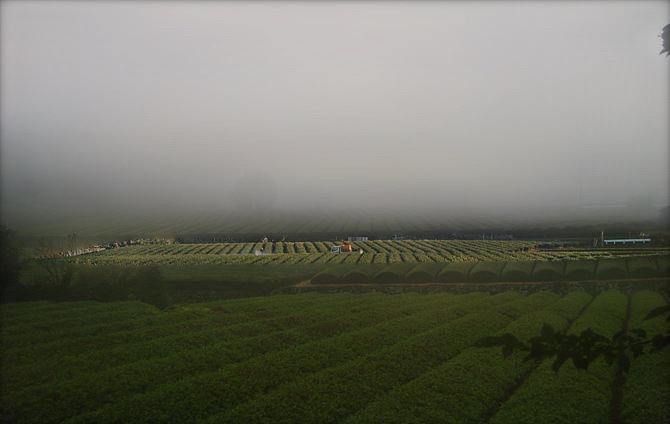}
\end{center}
       \caption{Results of our high-resolution image blending algorithm compared with baseline methods. From top to bottom: annotated object-level mask, composite copy-and-paste image, MPB, MSB, and GP-GAN.}
\label{fig:addtional_results_01}
\end{figure*}

\begin{figure*}
\begin{center}
    \includegraphics[height=0.18\linewidth]{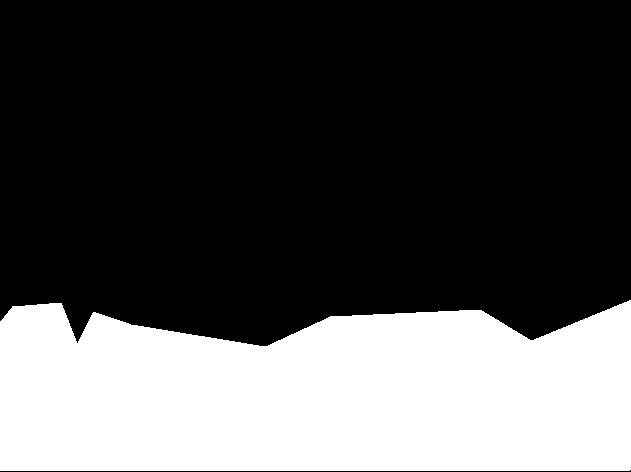}
    \includegraphics[height=0.18\linewidth]{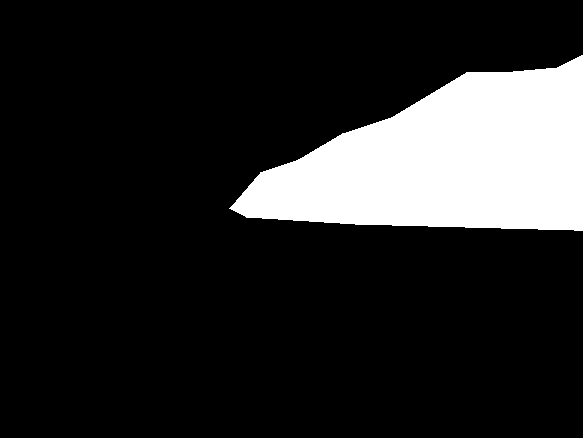}
    \includegraphics[height=0.18\linewidth]{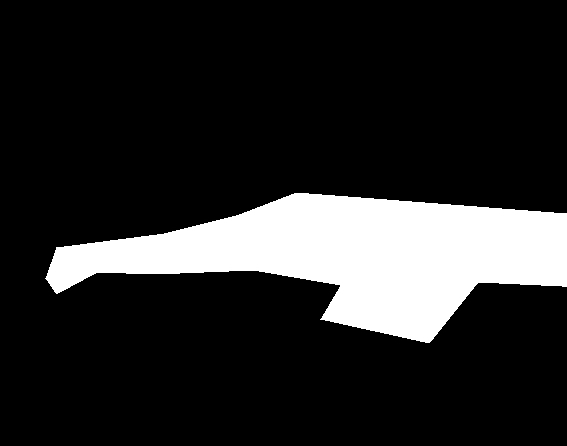}
    \includegraphics[height=0.18\linewidth]{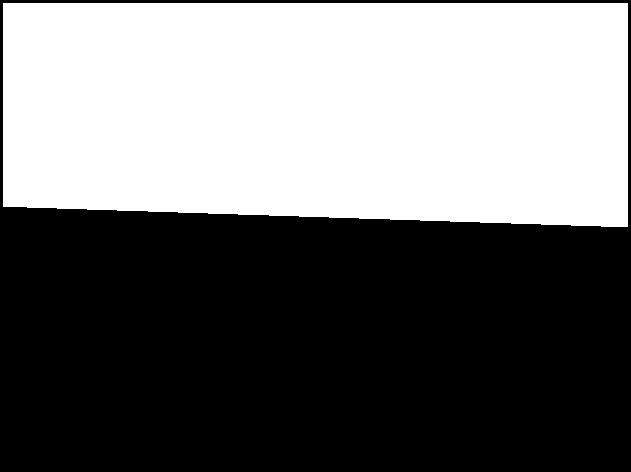}
    \\
    \includegraphics[height=0.18\linewidth]{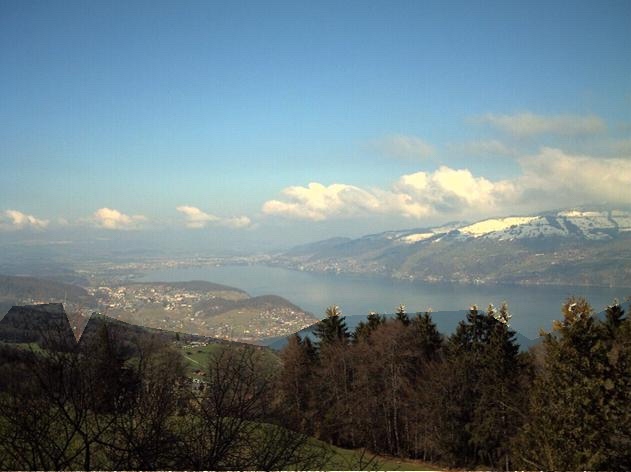}
    \includegraphics[height=0.18\linewidth]{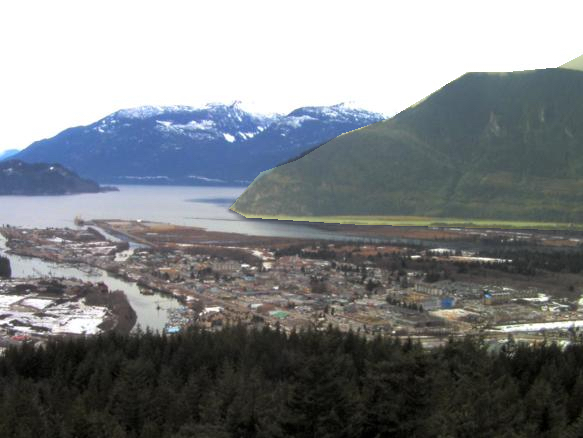}
    \includegraphics[height=0.18\linewidth]{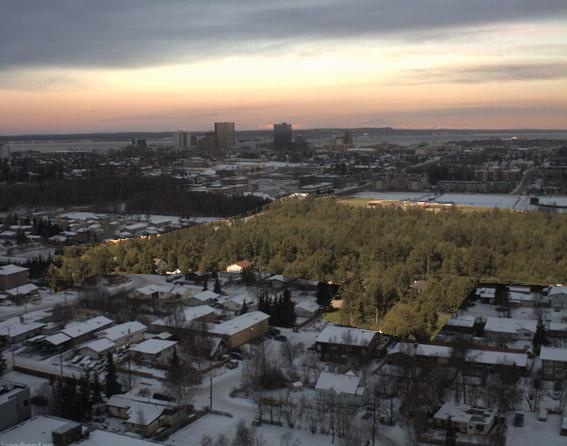}
    \includegraphics[height=0.18\linewidth]{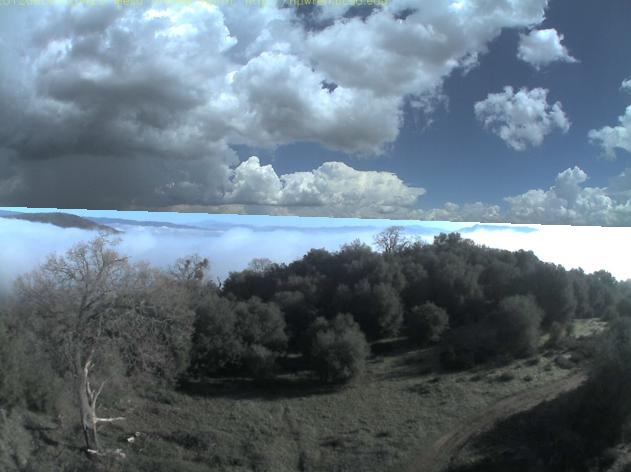}
    \\
    \includegraphics[height=0.18\linewidth]{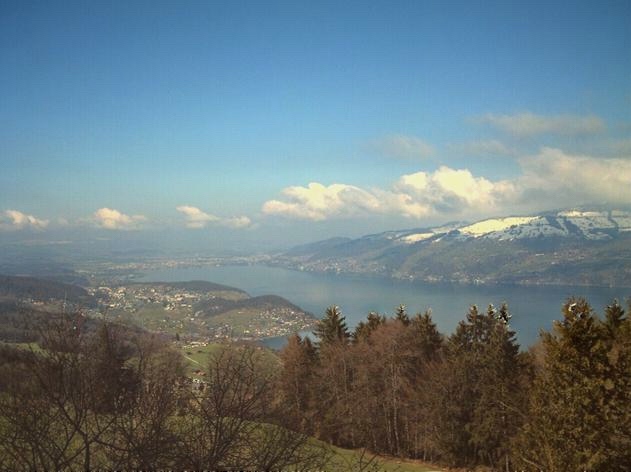}
    \includegraphics[height=0.18\linewidth]{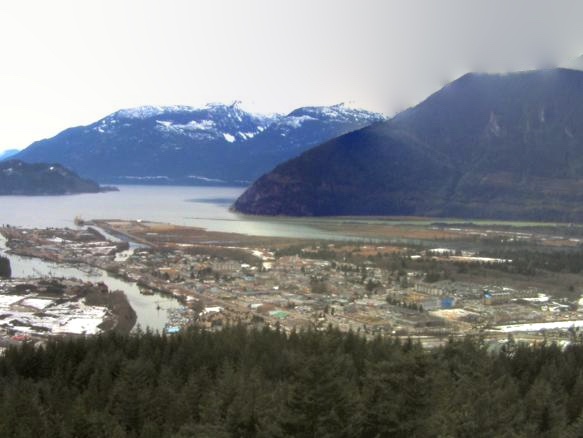}
    \includegraphics[height=0.18\linewidth]{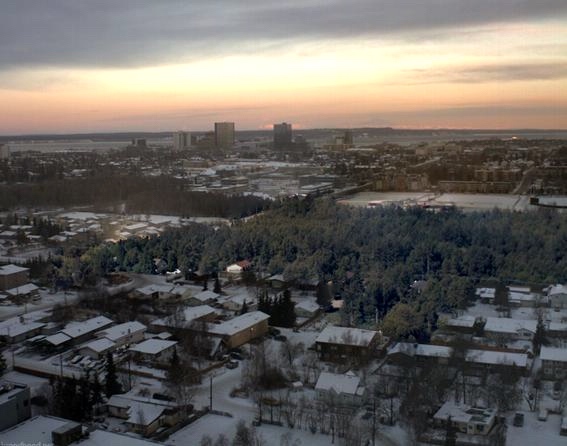}
    \includegraphics[height=0.18\linewidth]{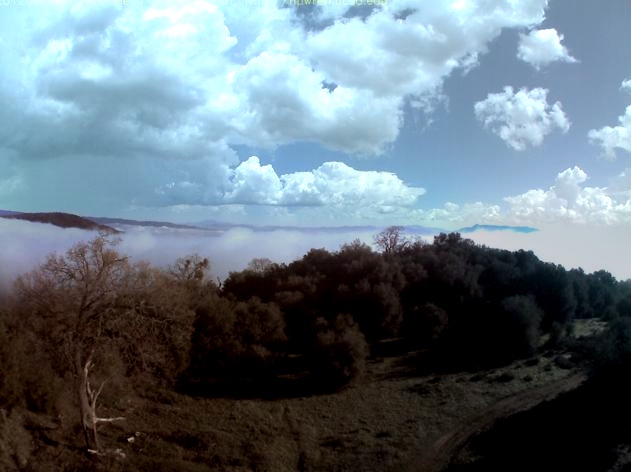}
    \\
    \includegraphics[height=0.18\linewidth]{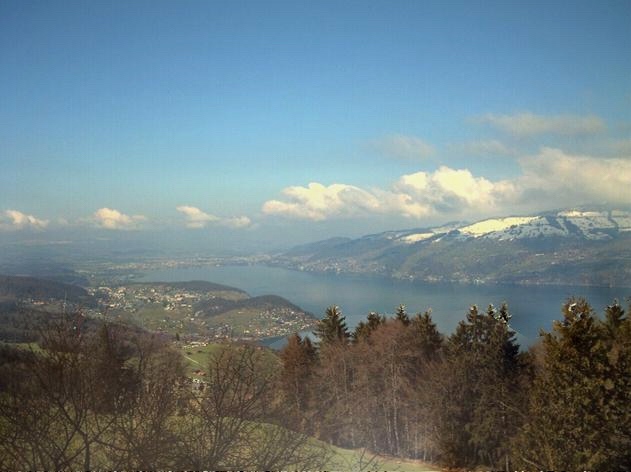}
    \includegraphics[height=0.18\linewidth]{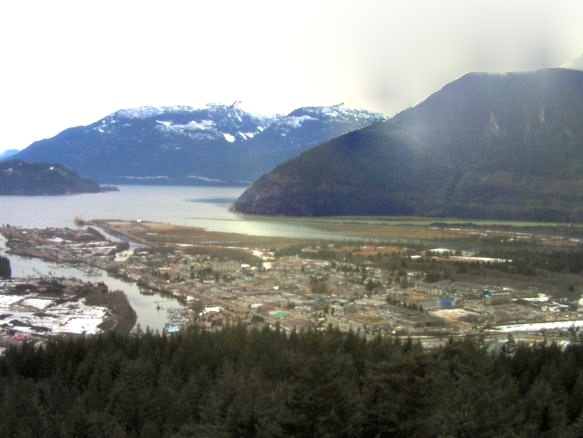}
    \includegraphics[height=0.18\linewidth]{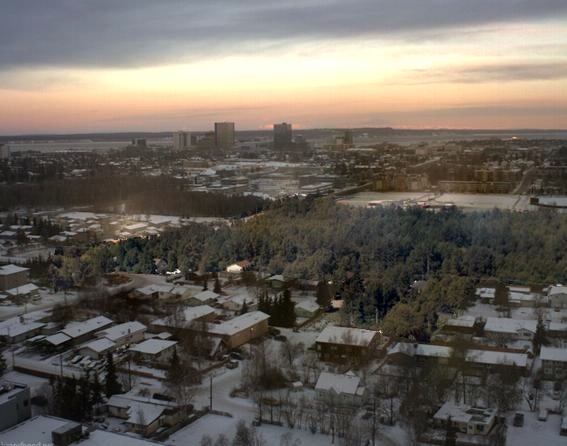}
    \includegraphics[height=0.18\linewidth]{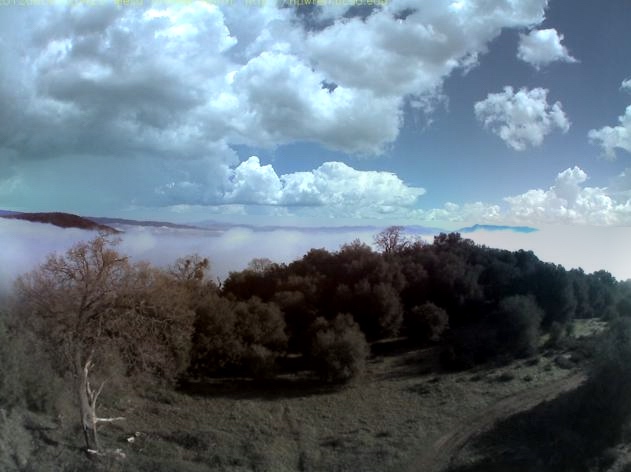}
    \\
    \includegraphics[height=0.18\linewidth]{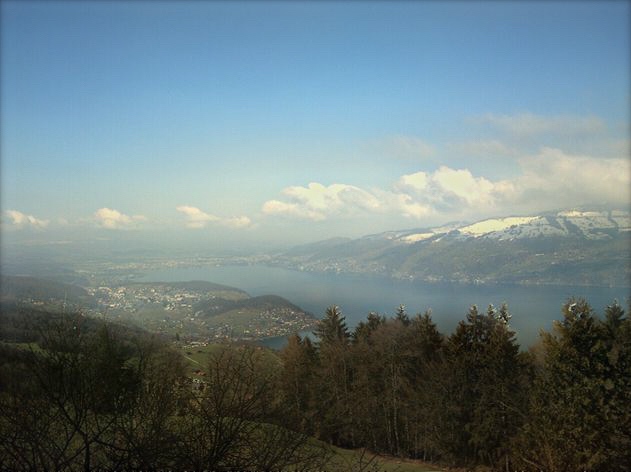}
    \includegraphics[height=0.18\linewidth]{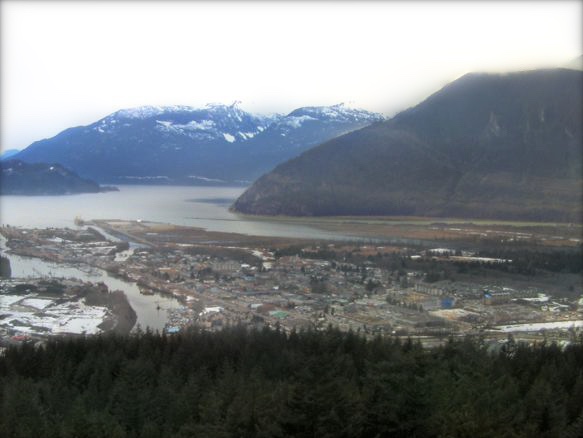}
    \includegraphics[height=0.18\linewidth]{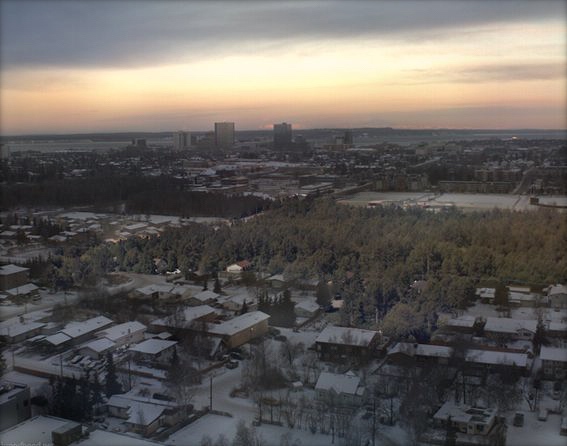}
    \includegraphics[height=0.18\linewidth]{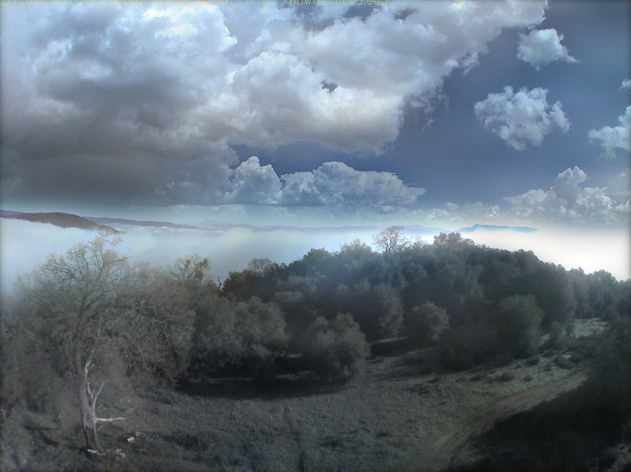}
\end{center}
       \caption{Results of our high-resolution image blending algorithm compared with baseline methods. From top to bottom: annotated object-level mask, composite copy-and-paste image, MPB, MSB, and GP-GAN.}
\label{fig:addtional_results_02}
\end{figure*}

More visual results are shown in Figure~\ref{fig:addtional_results_01} and Figure~\ref{fig:addtional_results_02}.

%

\end{document}